\PassOptionsToPackage{dvipsnames}{xcolor}
\documentclass[conference]{IEEEtran}

\let\labelindent\relax

\usepackage{times}
\usepackage[numbers, compress]{natbib}
\usepackage{multicol}
\usepackage[bookmarks=true]{hyperref}

\usepackage[utf8]{inputenc} 
\usepackage[T1]{fontenc}    
\usepackage{url}            
\usepackage{booktabs}       
\usepackage{amsfonts}       
\usepackage{nicefrac}       
\usepackage{microtype}      
\usepackage[dvipsnames]{xcolor}
\usepackage{graphicx} 
\usepackage{subcaption}
\usepackage{subfiles}

\graphicspath{{figs/}{../figs/}}

\usepackage[inline]{enumitem}

\newcommand{\green}[1]{\noindent\textcolor{ForestGreen}{#1}}

\usepackage{xspace}
\def\flow{RMP{flow}\xspace}
\def\algebra{RMP-algebra\xspace}
\def\tree{RMP-tree\xspace}

\def\pushforward{\texttt{pushforward}\xspace}
\def\pullback{\texttt{pullback}\xspace}
\def\resolve{\texttt{resolve}\xspace}

\def\forward#1{\texttt{task\_map(}#1\texttt{)}}
\def\rmp#1{\texttt{rmp\_eval(}#1\texttt{)}}

\def\jacobian#1#2{\texttt{jacobian(}#1\texttt{,}#2\texttt{)}}
\def\gradient#1#2{\texttt{gradient(}#1\texttt{,}#2\texttt{)}}
\def\jvp#1#2#3{\texttt{jvp(}#1\texttt{,}#2\texttt{,}#3\texttt{)}}
\def\identity#1{\texttt{copy(}#1\texttt{)}}
\def\algfull{RMPflow Reactive Motion Policy\xspace}
\def\alg{RMP$^2$\xspace}

\def\ad{automatic differentiation\xspace}
\def\AD{automatic differentiation\xspace}

\usepackage{amsmath}
\usepackage{amsfonts}
\usepackage{amssymb}
\usepackage{amsthm}
\usepackage{bm}
\usepackage{bbm}
\usepackage{mathtools}
\usepackage{enumitem}
\usepackage{thmtools,thm-restate}
\usepackage{algorithm}
\usepackage{algorithmic}
\usepackage{graphicx}
\usepackage{comment}
\usepackage[capitalise]{cleveref}

\theoremstyle{plain}

\theoremstyle{definition}

\theoremstyle{remark}

\def\CC{\mathcal{C}}

\def\LL{\mathcal{L}}
\def\MM{\mathcal{M}}

\def\TT{\mathcal{T}}
\def\VV{\mathcal{V}}

\def\Ab{\mathbf{A}}

\def\Jb{\mathbf{J}}
\def\Mb{\mathbf{M}}

\def\ab{\mathbf{a}}\def\cb{\mathbf{c}}
\def\fb{\mathbf{f}}
\def\gb{\mathbf{g}}

\def\ob{\mathbf{o}}
\def\qb{\mathbf{q}}
\def\ub{\mathbf{u}}
\def\vb{\mathbf{v}}\def\wb{\mathbf{w}}

\def\Rbb{\mathbb{R}}

\def\ett{\mathtt{e}}

\def\ltt{\mathtt{l}}

\def\rtt{\mathtt{r}}
\def\utt{\mathtt{u}}
\def\vtt{\mathtt{v}}

\def\R{\Rbb}

\def\t{\top}
\def\*{\star}

\newcommand{\norm}[1]{ \| #1 \|  }

\newcommand{\q}{\mathbf{q}}
\newcommand{\qd}{{\dot{\q}}}
\newcommand{\qdd}{{\ddot{\q}}}

\newcommand{\x}{\mathbf{x}}
\newcommand{\xd}{{\dot{\x}}}

\newcommand{\y}{\mathbf{y}}
\newcommand{\yd}{{\dot{\y}}}

\newcommand{\f}{\mathbf{f}}

\newcommand{\g}{\mathbf{g}}

\newcommand{\one}{\mathbf{1}}

\newcommand{\J}{\mathbf{J}}
\newcommand{\Jd}{{\dot{\J}}}

\newcommand{\M}{\mathbf{M}}

\newcommand{\sdot}[2]{\overset{\lower0.1em\hbox{$\scriptscriptstyle #2$}}{#1}}

\usepackage{pifont}

\newcommand{\cmark}{{\ding{51}}}%
\newcommand{\xmark}{\green{\ding{55}}}%
\newcommand{\algnote}[1]{{\it// #1}}

\title{\alg: A Structured Composable Policy Class for Robot Learning}

\author{Anqi~Li$^1$*, Ching-An~Cheng$^2$*, M.~Asif~Rana$^3$, Man Xie$^3$, Karl Van Wyk$^4$, Nathan~Ratliff$^4$, and Byron~Boots$^{1,4}$\\
  $^1$University of Washington, $^2$Microsoft Research, $^3$Georgia Institute of Technology, $^4$NVIDIA\\
  * equal contribution
  } 

\begin{document}
\maketitle


\begin{abstract}
    We consider the problem of learning motion policies for acceleration-based robotics systems with a structured policy class specified by \flow. \flow is a multi-task control framework that has been successfully applied in many robotics problems. Using \flow as a structured policy class in learning has several benefits, such as sufficient expressiveness, the flexibility to inject  different  levels of  prior  knowledge as well as the ability to transfer policies between robots. 
    However, implementing a system for end-to-end learning \flow policies faces several computational challenges. 
    In this work, we re-examine the message passing algorithm of \flow and propose a more efficient alternate algorithm, called \alg, that uses modern automatic differentiation tools (such as TensorFlow and PyTorch) to compute \flow policies.
    Our new design retains the strengths of \flow while bringing in advantages from automatic differentiation, including 
    {\it 1)} easy programming interfaces to designing complex transformations; 
    {\it 2)} support of general directed acyclic graph (DAG) transformation structures;
    {\it 3)} end-to-end differentiability for policy learning;
    {\it 4)} improved computational efficiency;.
    Because of these features, \alg can be treated as a structured policy class for efficient robot learning which is suitable encoding domain knowledge. 
    Our experiments show that using structured policy class given by \alg can improve policy performance and safety in reinforcement learning tasks for goal reaching in cluttered space. 
    
\end{abstract}

\section{Introduction} \label{sec:introduction}

Designing reactive motion policies is a fundamental problem in robotics. This problem has been tackled by analytic control techniques and machine learning tools, but they lead to different compromises. 
Traditionally control techniques have been used, offering motion policies with analytical forms and desirable properties such as stability, safety, and performance guarantees~\cite{bullo2005geometric,khalil2002nonlinear,nakanishi2008operational,siciliano2010robotics}. 
However, as tasks become more complex and the environment become less structured, synthesizing an analytical policy with such properties becomes difficult, and, even if one succeeds, the resulting policy may be highly sub-optimal~\cite{cheng2020rmpflow}.
%
In contrast to hand-designed control techniques, learning-based approaches (such as reinforcement learning~\cite{schulman2015high,schulman2017proximal} or imitation learning~\cite{ross2014reinforcement,ross2011reduction}) make minimal assumptions and promise to improve policy performance through interactions with the environment. 
However, such approaches often require many interactions to achieve reasonable results, especially when learning policies under sparse reward signals through reinforcement learning. 
Furthermore, most learning algorithms are sensitive to distribution shifts and have poor out-of-distribution generalization ability. 
For example, a policy that is trained to go to the left of a room in training often does not know how to go to the right, because such examples are never presented in training.

In practice, we desire motion policy optimization algorithms that possess \textit{both} the non-statistical guarantees from the control-based approaches and the flexibility of the learning-based approaches.
A promising direction is the use of \textit{structured policies}~\cite{choi2020reinforcement,chow2019lyapunov,dalal2018safe,meng2019NeuralAutoNavigation,mukadam2019riemannian}. Here the main idea is to apply learning to optimize only within policy parameterizations that have provable control-theoretic properties (such as stability and safety guarantees). In other words, it uses data to optimize the hyperparameters of a class of controllers with provable guarantees.
From a learning perspective, structured policies provide a way to inject prior knowledge about a problem domain into learning. More often than not, before running the robot in the field, we may know (approximate) kinematic and dynamic models, task constraints, and some notion of desired behaviors. For example, the kinematics of the robot is provided by the manufacturer, and we know generally that colliding with obstacles in the environment and hitting joint limits are undesirable. %
By using this information through a control framework to design structured policies, we can ensure policies produced a learning algorithm can always satisfy certain task specifications (such as safety) regardless how they are learned.

In this paper, we are interested in learning structured motion policies for acceleration-based robotics systems~\cite{siciliano2010robotics}. We adopt the \flow control framework~\citep{cheng2018rmpflow} as the foundation of the structured policy class.
\flow is an multi-task control algorithm that designs the acceleration-based motion policy by  decomposing a full robot control problem into smaller subtasks, such as goal reaching, collision avoidance, maintaining balance, etc. 
\flow treats these subtask spaces as manifolds and provides a message passing algorithm to combine subtasks policies into a stable motion policy for all the subtasks~\citep{cheng2018rmpflow}.
Because of its control-theoretic guarantees and computational efficiency, \flow has been applied to a range of robotic applications~\citep{2017_rss_system,li2019stable,li2019MultiAgentRMPsArXiv,meng2019NeuralAutoNavigation,paxton2019RLDS,sutanto2019TactileServoing,wingo2020extending}.

\begin{table*}[t]
    \small
    \centering
    \begin{tabular}{lllll}
      Algorithm   & Time  & Space  & Req. Tree  & Req. Auto.Diff.  \\
      \hline
     \alg (\cref{alg:new}) & \green{$O(Nbd^3)$} &  \green{$O(Ld^2 + Nd)$} & \xmark &  \cmark \\
     Na\"ive Implementation (\cref{alg:direct}) & $O(Nbd^3L)  $ & $O(NLd^2)$ & \xmark & \cmark  \\
     \flow (\cref{alg:rmpflow}) \citep{cheng2018rmpflow} & \green{$O(Nbd^3)$} & $O(Ld^2 + Nd^2)$ & \cmark & \xmark\footnotemark
    \end{tabular}
    \caption{Comparison between different implementations of the \flow policy for a graph/tree with $N$ nodes of dimension in $O(d)$, where $L\le N$ nodes are leaf nodes and the branching factor is in $O(b)$. Computing the task map is assumed to have time complexity of $O(Nbd^2)$ and memory complexity of $O(1)$. We assume the \ad library is based on reverse-mode \AD, for which calling the Gradient Oracle to compute the derivative of a scalar function with respect to a variable in $O(d)$ requires time complexity in $O(Nbd^2)$ and space complexity $O(Nd)$. See \cref{sec:method} for details. 
    } 
    \vspace{-3mm}
    \label{tb:theoretical comparison}
\end{table*}

Learning motion policies that can be expressed by the \flow framework, which we call the \textit{\flow policies}, has several advantages:
\begin{enumerate*}[label=\emph{\arabic*})]
    \item The task decomposition scheme in \flow provides an interface to inject different levels of prior knowledge.  As an example, one can hand design safety-critical subtask policies for, e.g. collision avoidance, and learn other subtask policies to improve the overall performance;  
    \item Hand-designed \flow policies have been applied to solve many real-world robotics applications that require complex motions, so \flow policies represent a sufficiently expressive for motion control problems;
    
    \item The \flow policy learned on one robot can be transferred to other robots because of its differential geometry centered design~\citep{cheng2018rmpflow}. This allows us to easily adopt existing hand-designed subtask policies in the literature to partially parameterize the \flow policy to help warm-start the learning process; 
\end{enumerate*}

Despite these promises, implementing a system for learning \flow policies faces several practical computational challenges, because the \flow message passing algorithm was originally designed for reactive control rather than learning,
One reason is that \flow uses a rather complicated user interface involving a tree data structure that is non-trivial for users to specify.
The other reason, which is perhaps more important, is that it is hard and computationally expensive to trace the gradient flow in the message passing algorithm of \flow, and yet tracing gradient flows is commonly required for end-to-end learning.
Although recent work has looked into learning with \flow~\citep{meng2019NeuralAutoNavigation,mukadam2019riemannian,rana2019learning}, these methods either largely simplify the learning problem so that differentiating through \flow is not needed~\citep{aljalbout2020learning,meng2019NeuralAutoNavigation,rana2019learning} or only allows for a very limited parameterization of the policy~\citep{mukadam2019riemannian}. For example, \citet{meng2019NeuralAutoNavigation} and \citet{rana2019learning} learn the subtask policies independently through imitation and then use \flow to combine the learned RMPs with other hand-specified policies; and \citet{mukadam2019riemannian} learn scalar weight functions for pre-defined subtask polices. Recently \citet{aljalbout2020learning} explored learning collision avoidance RMPs with \flow through reinforcement learning, where differentiating through \flow is not required as the gradients can be  estimated through samples and value function estimate. 

In this work, we propose a simple alternative algorithm, called \alg (\algfull), to replace the message passing algorithm in \flow for computing \flow policies, so that end-to-end learning \flow polices can be more easily implemented and scaled up in practice.
We emphasize that we \textit{do not} propose a new structured policy class, but a more efficient and flexible implementation of \flow policies. Policies realized by either the original message passing algorithm of \flow or our new \alg algorithm are therefore the same, and learning with them would lead to the same results statistically.
For clarity, we will refer to the \flow algorithm as the message passing routine in \flow (which our algorithm \alg replaces) and the \flow policy as the effective motion policy that \flow represents (which our \alg also represents).

\alg realizes the \flow policy by querying the Gradient Oracle~\citep{griewank2008evaluating} in an \AD library (such as TensorFlow~\citep{tensorflow2015-whitepaper} and PyTorch~\citep{paszke2017automatic}) instead of using the tree data structure and message passing steps of the \flow algorithm.
In comparison, \alg has several advantages over the original \flow algorithm: 
\begin{enumerate*}[label=\emph{\arabic*})]
    \item \alg allows for a simper user interface. The user only needs to specify the task map using an \ad library, the automatically constructed (directed acyclic) computational graph can then be used for computing the \flow policy. 
    \item \alg relaxes the assumption on using tree-structured task maps in the \flow algorithm~\citep{cheng2018rmpflow} to work with \emph{any} directed acyclic graph (DAG) task maps. 
    \item \alg is much easier to implement in conjunction with learning algorithms: as \alg is implemented using operators supported by \ad libraries, it is convenient to take the gradient, or higher order derivatives, of \emph{any function} with respect to the parameters in subtask policies and task maps. 
    \item \alg uses a smaller memory footprint than \flow, while having the same time complexity (see \cref{tb:theoretical comparison}).
\end{enumerate*}

These computational advantages make \alg  generally applicable to many end-to-end learning scenarios and algorithms. In the rest of the paper, we provide the details of our new algorithmic design and its complexity analysis. At the end of the paper, we validate our algorithm \alg in applications of learning acceleration-based motion control policies with reinforcement learning in simulated reaching tasks with a three-link robot arm and a Franka robot arm.

\vspace{-0.5mm}
\section{Background} \label{sec:background}
\vspace{-0.5mm}
\subsection{Acceleration-based Motion Control}

We focus on learning policies for controlling the motion of acceleration-based robotics systems. Typically this type of kinematic control problems  arises when one wishes to reactively generate smooth reference trajectories for a low-level tracking controller, or wants to control a robot that is fully actuated and feedback linearized (e.g., by an inverse dynamics model).

Suppose the robot's configuration space $\CC$ (e.g. the joint space of a robot with revolute joints) is a $d$-dimensional smooth manifold that can be described by generalized coordinates $\q \in \R^d$.
We can view the acceleration-based motion control problem as a continuous-time deterministic Markov decision process (MDP): the state is the position-velocity $(\q, \qd)$, the action is the acceleration $\qdd$, the transition is the integration rule, and the reward is defined to encourage desired behaviors for a task (such as smooth, collision-free motions).
Our goal is to find an acceleration-based policy $\pi$ such that the system following $\qdd=\pi(\q, \qd)$ would exhibit good performance for the tasks of interest. 

In these motion control problems, the desired behavior of a task (equivalently the reward function) is often not directly described in the generalized coordinates $\qb \in \R^d$, but in terms of another set of task coordinates $\x\in\R^m$ that are related to the generalized coordinates through a nonlinear mapping $\psi$, i.e. $\x = \psi(\q)$.
For example, in controlling an anthropomorphic robot, we wish to control the torso's motion to maintain the stability and the reachable region of the hands, whose performance is more easily described in the coordinates of the torso and the hand rather than directly in the joint space (i.e. the configuration space). 
We call this mapping $\psi$ the \emph{task map} and refer to the image manifold of $\CC$ under $\psi$ as the \emph{task space}, which is denoted as $\TT$. 
As shown in the above example, the task here is often multi-task in nature requiring the robot to satisfy various performance criteria. 
Mathematically, this implies that the overall task space $\TT$ is a composition of many subtask spaces, such as $\TT = \prod_{k=1}^K \TT_k$ in a $K$-task control problem based on subtask manifolds $\{\TT_k\}_{k=1}^K$. 
These subtask coordinates are often not independent but intertwined together as the image of the common configuration space $\CC$ under the task map $\psi$ (in the previous example, moving the torso affects the position of the robot hand). Therefore, generally, policies designed for each subtask cannot be trivially combined together (e.g., with using a convex combination) to generate a good policy for the full task.

\vspace{-0.5mm}
\subsection{\flow: A Framework for Multi-Task Problems}
\vspace{-0.5mm}

\flow~\citep{cheng2018rmpflow} is a control-theoretic computational framework designed to address the multi-task control problems mentioned above.
Inspired by the geometric control theory~\citep{bullo2005geometric}, \flow resolves the conflicts between different subtasks by describing each subtask policy as a Riemannian Motion Policy (RMP)~\citep{ratliff2018riemannian}. An RMP associates a subtask (e.g. for the $k$th subtask) not only with the desired acceleration $\ab_k^d(\x,\xd)$, but also with a positive semi-definite matrix function $\M_k(\x,\xd)$ that depends on the state (i.e. the position and the velocity) of the subtask. 
Given RMPs for the subtasks, \flow generates the policy $\pi$ on the configuration space by combining these subtask RMPs through message passing on a tree data structure of manifolds (called the \tree), where the root and leaf nodes correspond to the configuration space $\CC$ and the subtask spaces $\{\TT_k\}$, and an edge represents a smooth map from a parent node manifold to its child node manifold.

This message passing scheme of \flow effectively realizes a differential-geometric operation, called the \emph{pullback}, which propagates differential forms from the subtask manifolds to the configuration space manifold.
As a result, it can be proved that the final policy $\pi$ output by \flow is Lyapunov stable, when $\Mb_k$ is derived appropriately from a Riemannian metric that describes the motion induced by $\ab_k^d$ for each policy~\citep{cheng2018rmpflow,li2019stable}. 
These nice control-theoretical properties makes \flow a promising candidate of the structured policy class for acceleration-based motion control.

Let us provide some intuitions as to why \flow works. 
Here we take an optimization viewpoint recently made in~\citep{cheng2020efficient} (rather than the common geometric control viewpoint of \flow used in the literature) and show the optimization problem that \flow effectively solves in the pullback operation when generating the multi-task control policy. This insight explains the properties of \flow, which is more directly related to our proposed algorithm \alg, without going through its complex algorithmic steps. The message passing procedure of \flow is listed in~\cref{alg:rmpflow} and a detailed description can be found in~Appendix~\ref{sec:rmpflow_alg}.

\begin{algorithm}[tb]
    \small
    \begin{algorithmic}[1]
    \STATE \textbf{Input: } root state $(\q, \qd)$, \tree \texttt{T}, RMPs \texttt{rmp\_eval}
    \STATE \textbf{Return: } motion policy $\pi(\q, \qd)$
    \STATE $\texttt{nodes} \gets$ \texttt{T}.\texttt{topologically\_sorted\_nodes()}\\
    \vspace{1mm}
           {\it// forward pass }
    \STATE \textbf{For} \texttt{node} \textbf{in} \texttt{nodes}: \algnote{from root to leaves}
    \STATE  \qquad \textbf{For} \texttt{child} \textbf{in} $\texttt{node}.\texttt{children}$:
    \STATE \qquad \qquad $\texttt{child}.\texttt{state} \gets \pushforward(\texttt{node}.\texttt{state})$\\
    \vspace{1mm}
         {\it// evaluate leaf RMPs}
    \STATE \textbf{For} \texttt{node} \textbf{in} \texttt{T}.\texttt{leaves}:
    \STATE \qquad $\texttt{node}.\texttt{rmp} \gets \rmp{\texttt{node}.\texttt{state}}$ \\
    \vspace{1mm}
         {\it// backward pass}
    \STATE \textbf{For} \texttt{node} \textbf{in} \texttt{reversed}(\texttt{nodes}): \algnote{from leaves to root }
    \STATE \qquad  \texttt{node.rmp} $\gets$ \pullback(\texttt{node.child.rmps})\\
        {\it// resolve for the motion policy}
    \STATE $\pi(\q, \qd) \gets$ \resolve(\texttt{T.root.rmp})\\
    \end{algorithmic}
    \caption{The \flow Algorithm (Message Passing) \citep{cheng2018rmpflow}}
    \label{alg:rmpflow}
\end{algorithm}

Consider an RMP-tree with a set of nodes $\VV$. Let $\LL:=\{\ltt_k\}_{k=1}^K\subset\VV$ be the set of leaf nodes and $\rtt$ be the root node. 
\citet[Chapter 11.7]{cheng2020efficient} observed that there is a connection between the message passing algorithm of \flow and sparse linear solvers: the desired acceleration of the \flow policy $\pi(\q,\qd)$ is the solution to the following least squares problem, and the message passing routine of \flow effectively uses the duality of \eqref{eq:rmpflow_objective} and the sparsity in the task map to efficiently compute its solution. 
\begin{align}
      \min_{\{\ab_\vtt: \vtt\in\VV\}}&   \quad \sum_{k=1}^K\,\frac12\,\norm{\ab_{\ltt_k} - \ab_k^d}_{\M_k}^2,\label{eq:rmpflow_objective}\\
        \textrm{s.t.} &\quad\hspace{1mm} \ab_\vtt = \J_{\vtt;\utt} \ab_\utt + \Jd_{\vtt;\utt}\,\xd_\utt,\label{eq:rmpflow_constraint}\\
        & \qquad\forall\,\vtt\in\VV\setminus \rtt, \quad\utt = \texttt{parent}(\vtt)\nonumber
\end{align}
where, for a leaf node $\ltt_k$, $\ab_k^d$ and $\M_k$ together are a leaf-node RMP, $\J_{\vtt;\utt}$ denotes the Jacobian of the task map from node $\utt$ to node $\vtt$, and $\Jd_{\vtt;\utt}$ is the time-derivative of Jacobian $\J_{\vtt;\utt}$. The objective in~\eqref{eq:rmpflow_objective} is the sum of deviation between the acceleration $\ab_{\ltt_k}$ on a leaf space and the desired one, $\ab_k^d$, weighted by the importance matrix $\Mb_k $. The constraints~\eqref{eq:rmpflow_constraint} enforce the accelerations to be consistent with the maps between spaces. 
In other words, the leaf-node RMPs in \flow defines a trade-off between different subtask control schemes and the policy of \flow is the optimal solution that can be realized under the geometric constraints between the subtask spaces and the configuration space.

Implementing a system for learning \flow policies, however, can be computationally challenging, because the user needs to implement the complex message passing algorithm and the data structure described in~Appendix~\ref{sec:rmpflow_alg} (which we omitted here due to its complexity!). Moreover, the user may need to trace the gradient flow through this large computational graph.
This difficulty has limited the applicability of existing work of end-to-end learning of \flow policies 
\cite{meng2019NeuralAutoNavigation,mukadam2019riemannian,rana2019learning,rana2020towards}. Improving the efficiency and simplicity of implementing \flow policies is the main objective of our paper.

\vspace{-0.5mm}
\section{\alg based on Automatic Differentiation} \label{sec:method}
\vspace{-0.5mm}

We propose  an alternate algorithm to implement the \flow policy. 
Our new algorithm, \alg,  works by properly calling the basic oracles (such as evaluation and the Gradient Oracle) of an \ad library, without using the message passing algorithm of \flow in Appendix~\ref{sec:rmpflow_alg}.
The result is an easy-to-use and computationally efficient framework suitable for end-to-end learning \flow policies.

\subsection{Key Idea} 

We observe that the constrained optimization problem~\eqref{eq:rmpflow_objective}--\eqref{eq:rmpflow_constraint} is equivalent to the following unconstrained least-squares optimization problem if the constraints are aggregated: 
\begin{equation}\label{eq:unconstrained LS formulation}
    \min_{\ab_\rtt' \in \R^d}  \quad \sum_{k=1}^K\,\frac12\,\norm{ \Jb_{\ltt_k;\rtt} \ab_\rtt' + \Jd_{\ltt_k;\rtt}\qd  - \ab_k^d}_{\M_k}^2,
\end{equation}
where
$\Jb_{\ltt_k;\rtt}$ is the Jacobian matrix of the subtask map from the root node $\rtt$ to the $k$th leaf node $\ltt_k$. Note that the Jacobians and velocities here are treated as constants in the optimization as they are only dependent on the state (not the accelerations). 

This observation implies that we can compute the \flow policy by solving~\eqref{eq:unconstrained LS formulation}, which has the following closed-form solution:
\begin{align} \label{eq:closed-form solution}
    \ab_\rtt = \underbrace{ \left( \sum_{k=1}^K \Jb_{\ltt_k;\rtt}^\t \M_k \Jb_{\ltt_k;\rtt} \right)^{\dagger} }_{\Mb_\rtt^\dagger}
    \underbrace{  \left( \sum_{k=1}^K \Jb_{\ltt_k;\rtt}^\t  \M_k  (\ab_k^d - \dot{\Jb}_{\ltt_k;\rtt}\qd)   \right) }_{\fb_\rtt}.
\end{align}
This means that if we can compute the solution in \eqref{eq:closed-form solution} efficiently for a large set of sparsely-connected manifolds, then we can realize the \flow Policy without the \flow algorithm (in \cref{alg:rmpflow}).

\subsection{\alg based on Reverse Accumulation}

We propose \alg (\cref{alg:new}), an efficient technique that computes the closed-form solution~\eqref{eq:closed-form solution} using automatic differentiable libraries, which are commonly used in modern machine learning pipelines.  
As we will show in Appendix~\ref{sec:complexity},
\alg has the same time complexity of as \flow~\cite{cheng2018rmpflow} while enjoying a smaller memory footprint. More importantly, \alg provides a more simple and intuitive interface for learning.

In \AD libraries, a {computation graph} is automatically built as transform maps are specified, and derivatives are computed through message passing on the computation graph. \alg leverages this feature so that no additional data structure and message passing routine are needed, whereas \flow requires the user to specify an \tree and implement the message passing algorithm  on it. 

\alg uses the following common functionalities provided by \AD libraries:
\begin{itemize}
    \item \gradient{$s$}{$\ub$}: the gradient operator. It computes the gradient of scalar graph output $s\in\R$ with respect to graph input vector $\ub\in\R^n$ through back-propagation; 
    
    \item \jacobian{$\vb$}{$\ub$}: the Jacobian operator. It computes the Jacobian matrix $\partial_\ub \vb\in\R^{m\times n}$  through back-propagation; 
    \texttt{jacobian} is equivalent to multiple calls of \texttt{gradient}.
    
    \item $\jvp{\vb}{\ub}{\wb}$: the Jacobian-vector product. It computes $\left(\partial_\ub \vb\right) \wb\in\R^m$ 
    given graph input vector $\ub\in\R^n$, graph output vector $\vb\in\R^m$, and an addition vector $\wb\in\R^n$. 
    The Jacobian-vector product can be efficiently realized by \texttt{gradient}
    (see Algorithm~\ref{alg:jvp}) using a technique, called reverse accumulation~\citep{christianson1992automatic} (also known as double backward). The algorithm computes Jacobian-vector product through $2$ passes. An auxiliary all ones vector $\bm\lambda$ is created and the gradient with respect to $\bm\lambda$ is tracked (line~\ref{ln:dummy}). 
\end{itemize}

\begin{algorithm}[t]
    \begin{algorithmic}[1]
    \small
    \STATE \textbf{Input: } $\ub$, $\vb$, $\wb$
    \STATE \textbf{Return: $\left(\partial_\ub \vb\right) \wb$}\label{ln:dummy}
    \STATE $\bm\lambda \gets \one$ \algnote{dummy variable for reverse accumulation}
    \STATE $\gb\gets \gradient{\bm\lambda^\t \vb}{\ub}$ \algnote{sum of partial derivatives}\label{ln:auxiliary}
    \STATE compute Jacobian-vector product \[\left(\partial_\ub \vb\right)\wb\gets\gradient{\g^\t\vb}{\bm\lambda}\]\label{ln:jvp}
    \vspace{-5mm}
    \end{algorithmic}
    \caption{Jacobian-vector product~\cite{christianson1992automatic} $\jvp{\vb}{\ub}{\wb}$} \label{alg:jvp}
\end{algorithm}

\begin{algorithm}[t]
    \small
    \begin{algorithmic}[1]
    \STATE \textbf{Input: } root state $(\q, \qd)$, \texttt{task\_map}, \texttt{rmp\_eval}
    \STATE \textbf{Return: } motion policy $\pi(\q, \qd)$ \\
    \vspace{1mm}
    \algnote{forward pass}
    \STATE $\{\x_k\}_{k=1}^K\gets\forward{\q}$ \label{ln:taskmap}
    \vspace{1mm}
    \STATE $\{\xd_k\}_{k=1}^K\gets\jvp{\{\x_k\}_{k=1}^K}{\q}{\qd}$ \algnote{leaf node velocity}\label{ln:velocity}
    \vspace{1mm}
    \STATE $\{\cb_k\}_{k=1}^K\gets\jvp{\{\xd_k\}_{k=1}^K}{\q}{\qd}$  \algnote{curvature terms }
    \label{ln:curvature}\\
    \vspace{2mm}
    \algnote{evaluate leaf RMPs}
    \vspace{1mm}
    \STATE $\{(\M_k,\ab_k^d)\}_{k=1}^K\gets \rmp{\{(\x_k,\xd_k)\}_{k=1}^K}$\\
    \vspace{2mm}
    \algnote{backward pass}
    \vspace{1mm}
    \STATE  $\q'= \identity{\q}$, $\q''= \identity{\q}$ \algnote{copies without gradient}
    \vspace{1mm}
    \STATE  $\{\x'_k\}_{k=1}^K\gets\forward{\q'}$,  $\{\x''_k\}_{k=1}^K\gets\forward{\q''}$ \algnote{mirrored images}
    \vspace{1mm}
    \STATE  $r\gets\sum_{k=1}^K (\x_k')^\t \M_k \x''_k$,\quad $s\gets \sum_{k=1}^K(\x'_k)^\t\M_k (\ab_k^d - \cb_k)$ \algnote{auxiliary variables }\label{ln:auxiliary uw}
    \vspace{1mm}
    \STATE $\M_\rtt \gets \jacobian{\gradient{r}{\q}}{\q'}$,\quad  \algnote{root matrix}\label{ln:backward2} \\
    \STATE  $\fb_\rtt\gets \gradient{s}{\q}$ \algnote{root force}\label{ln:backward1}\\
    \vspace{1mm}
    \algnote{resolve for the motion policy}
    \STATE $\pi(\q, \qd) \gets \M_\rtt^{\dagger}\,\f_\rtt$
    \end{algorithmic}
    \caption{\alg} \label{alg:new}
\end{algorithm}

By using the \texttt{gradient}, \texttt{jacobian}, \texttt{jvp} operators, \alg in \cref{alg:new} efficiently can compute  quantities needed for evaluating~\eqref{eq:closed-form solution}. 
In the forward pass of \alg (\cref{alg:new}, line 3--5), \alg evaluate subtask maps and compute their velocities and curvature terms using Jacobian-vector products. 
Specifically, for the $k$-th subtask map $\psi_{\ltt_k;\rtt}:\q\mapsto \x_k$ (which we may think as the map from the joint space of a robot manipulator to the workspace), the velocity $\xd_k:=\J_{\ltt_k;\rtt}\qd$ and curvature term $\cb_k:=\Jd_{\ltt_k;\rtt}\qd$ on the subtask space can be computed through Jacobian-vector products:
\begin{equation}
        \xd_k= \jvp{\x_k}{\q}{\qd},\quad \text{and} \quad \cb_k = \jvp{\xd_k}{\q}{\qd}. 
\end{equation}
Using $\{(\x_k, \xd_k)\}$, \alg then evaluates the values of the leaf RMPs (line 6), which implicitly define the objective of the weighted least squares problem in \eqref{eq:unconstrained LS formulation}.
Next, in the backward pass (line 7--11), \alg computes the pullback force $\fb_\rtt$ and importance weight matrix $\Mb_\rtt$ in~\eqref{eq:closed-form solution} using reverse accumulation (i.e., the technique used in \texttt{jvp} in \cref{alg:jvp}).
This is accomplished by creating auxiliary variables $\q'$ and $\q''$ (which have the same numerical value as $\q$ but are different nodes in the computational graph of \AD) and their mirrored task images (line 7 and 8), and then querying the gradients and Jacobians: 
\begin{equation}
    \small
    \begin{split}
        \fb_\rtt &= \texttt{gradient}\Big(\sum_{k=1}^K(\x'_k)^\t\M_k (\ab_k^d - \cb_k)\texttt{,} \q'\Big),\\
        \Mb_\rtt &= \texttt{jacobian}\Big(\texttt{gradient}\big(\sum_{k=1}^K (\x_k')^\t \M_k \x_k''\texttt{,}\q'\big)\texttt{,}\q''\Big),\\
    \end{split}
\end{equation}
where $\x'_k = \psi_k(\q')$, $\x''_k = \psi_k(\q'')$, and $\q=\q'=\q''$.

\subsection{Complexity of \alg}\label{sec:short complexity}

In Appendix~\ref{sec:complexity}, we analyze the time and space complexities of \alg. We show that \alg has a time complexity of $O(Nbd^3)$ and a memory complexity of $O(Nd+Ld^2)$, where $N$ is the total number of nodes, $L$ is the number of leaf nodes, $b$ is the maximum branching factor, and $d$ is the maximum dimension of nodes. 
In comparison, we prove in Appendix~\ref{app:complexity} that the original \flow algorithm (\cref{alg:rmpflow}) by \citep{cheng2018rmpflow} has a time complexity of $O(Nbd^3)$ and a \emph{worse} space complexity of $O(Nd^2+Ld^2)$. 
Please see \cref{tb:theoretical comparison} for a summary.

\subsection{Discussion: A Na\"ive Alternative Algorithm}

With the \flow policy expression in \eqref{eq:closed-form solution}, 
one may attempt to explicitly compute the matrices and vectors listed in \eqref{eq:closed-form solution} using \AD and then combine them to compute the \flow policy.
This idea leads to a conceptually simpler algorithm, shown in~\cref{alg:direct}, which  directly computes the Jacobians for the task maps through the \texttt{jacobian} operator provided by the \AD library, and then compute the root RMPs $\Mb_\rtt$ and $\fb_\rtt$ based on \eqref{eq:closed-form solution}.

However, this na\"ive  approach turns out to be not as efficient as \alg and \flow. Because the Jacobian matrix $\Jb_k:=\Jb_{\ltt_k;\rtt}$ is constructed here (line~\ref{ln:jacobian_jacobian} in \cref{alg:direct}), the time complexity of the na\"ive algorithm is $O(L)$ times larger than \alg, where $L$ is the number of the leaf nodes. For a binary tree with $N$ nodes, this means the time complexity is in $O(N^2)$, not the $O(N)$ of \alg and \flow. 

Moreover, this na\"ive algorithm also has a large space complexity of $O(NLd^2)$. This large space usage is created by the computation of the curvature term: the curvature term here is computed through differentiating velocities $\{\xd_k\}$ that are computed by the explicit Jacobian vector product in line~\ref{ln:jacobian_velocity}; as a result, the intermediate graph created by the Jacobian needs to be stored, which is the source of high memory usage. A simple to fix this memory usage is to compute the curvature term instead by two calls of Jacobian-vector-product (as in \alg). This modified algorithm has the time complexity of $O(Nbd^3L)$ (still $O(L)$ times larger than \alg and \flow) but an space complexity of $O(Nd+Ld^2)$ (same as \alg). 

\begin{algorithm}[t]
    \small
    \begin{algorithmic}[1]
    \STATE \textbf{Input: } root state $(\q, \qd)$, \texttt{task\_map}, \texttt{rmp\_eval}
    \STATE \textbf{Return: } motion policy $\pi(\q, \qd)$ \\
    \vspace{1mm}
    \algnote{forward pass}
    \STATE $\{\x_k\}_{k=1}^K\gets\forward{\q}$\label{ln:jacobian_taskmap} 
    \vspace{1mm}
    \STATE $\{\J_k\}_{k=1}^K\gets\jacobian{\{\x_k\}_{k=1}^K}{\q}$\label{ln:jacobian_jacobian} 
    \vspace{1mm}
    \STATE $\{\xd_k\}_{k=1}^K\gets\{\J_k \qd_k\}_{k=1}^K$ \algnote{leaf space velocity}\label{ln:jacobian_velocity}
    \vspace{1mm}
    \STATE $\{\cb_k\}_{k=1}^K\gets\jvp{\{\xd_k\}_{k=1}^K}{\q}{\qd}$\algnote{curvature terms}\label{ln:jacobian_cvp} \\
    \vspace{1mm}
    \algnote{evaluate leaf RMPs}
    \STATE compute leaf RMPs \[\{(\M_k,\ab_k^d)\}_{k=1}^K\gets \rmp{\{(\x_k,\xd_k)\}_{k=1}^K}\]\\
    \algnote{backward pass}
    \STATE compute root RMP \[\textstyle \M_\rtt \gets \sum_{k=1}^K \J_k^\t \M_k \J_k,\quad \f_\rtt\gets \sum_{k=1}^K \J_k^\t\M_k(\ab_k^d - \cb_k)\] \label{ln:jacobian_backward} \\
    \vspace{-4mm}
    \algnote{resolve for the motion policy}
    \STATE $\pi(\q, \qd) \gets \M_\rtt^{\dagger}\,\f_\rtt$
    \end{algorithmic}
    \caption{A Na\"ive Implementation} \label{alg:direct}
\end{algorithm}

\vspace{-3mm}
\subsection{Key Benefits of \alg}


\textbf{Simpler User Interface: }
The major benefit of \alg is that it is much easier to implement and apply than \flow, while producing the same policy and having the same time complexity. For \alg, the user no longer needs to construct the \tree data structure (just like implementing a neural network architecture from scratch) or implement the message passing algorithm. Instead, the user only needs to specify the task maps with \AD libraries, and the policy can be computed through standard operators in \AD libraries. See Appendix~\ref{sec:interface_case_study} for a case study.

\textbf{More General Taskmaps: }
Another benefit of \alg is that it supports task maps described by \emph{arbitrary} directed acyclic graphs (DAGs), whereas \flow is limited to tree-structured task maps. While \citet{cheng2018rmpflow} show that every task map has a tree representation, not all motion control problems have an \emph{intuitive} \tree representation (e.g. multi-robot control~\cite{li2019MultiAgentRMPsArXiv}). 
If they are implemented using a tree structure, extra high-dimensional nodes would be induced and the user interface becomes tedious.

\textbf{Differentiable Policies: }
Since \alg is implemented using the \ad libraries, computational graph can be automatically constructed while calculating the policy. This allows for convenient gradient calculation of \emph{any function} with respect to \emph{any parameters} in, e.g., parameters used to describe task maps and RMPs. This fully differentiable structure is useful to end-to-end learning of these parameters in many scenarios. 

\textbf{Smaller Memory Footprint: }
As is analyzed in~Appendix~\ref{sec:complexity}, \alg has a memory footprint of $O(Nd+Ld^2)$, which is smaller than \flow (see \cref{tb:theoretical comparison}).

In summary, \alg is an efficient algorithm that is easier to implement and apply, while providing a more convenient interface for learning applications. 

\vspace{-0.5mm}
\section{\alg for Learning}\label{sec:learning}
\vspace{-0.5mm}

In this section, we discuss various options of using \alg to parameterize structured policies for learning. We note that some examples below have already been explored by existing work using the \flow algorithm. However, in most cases, realizing these ideas with the new  \alg algorithm instead of the origin message passing routine of the \flow algorithm would largely simplify the setup and implementation, as \alg provides a more natural interface for learning applications. Moreover, \alg enables graph-structured policies, whereas the \flow algorithm works only with tree-structured policies.

\subsection{Parameterizing \alg Policies}

\alg policies are alternate parameterizations of \flow policies. They differ only in the way how \eqref{eq:closed-form solution} is computed (\alg policies use \AD whereas \flow policies use the message passing routine in the \flow algorithm). These two paramertizations therefore have the same representation power, but potentially the \alg poclies are more computationally efficient, because the tree structure used in \flow may not be the most natural way to describe the task relationship.

\textbf{Learnable leaf RMPs: } One way to parameterize \alg policies is through parameterizing leaf RMPs in \alg. There have been existing work exploring various ways to represent RMPs with neural networks so that the resulting policy can have certain theoretical properties, e.g. positive-definite weight matrices, Lyapunov-type stability guarantees, etc.~\cite{mukadam2019riemannian,rana2019learning}

\textbf{Learnable task maps: } Perhaps less obviously, one can also learn the task maps. For example, existing work has developed task map learning techniques such that the latent space policies take in simple forms or are easier to be learned~\cite{rana2020euclideanizing,urain2020imitationflow}. These representation learning techniques can be easily realized by \alg for learning \alg polices. 

\subsection{Learning Setups}

Because \alg is based on \AD, it can be implemented naturally with the typical machine learning pipelines. Below we discuss  common scenarios.

\textbf{Supervised Learning: } 
When there is an expert policy, one common scenario for robot learning is behavior cloning~\citep{pomerleau1988alvinn}, which minimizes the empirical difference between the learner and expert policies on a dataset collected by the expert policy. Gradient-based algorithms are often used to minimize the error, which requires computing the derivative of the acceleration-based policy with respect to the parameters. Due to the difficulty in differentiating through the \flow algorithm, most existing work on learning the RMPs in \flow policies with supervised learning either differentiates through an approximate algorithm (e.g. without the curvature terms)~\cite{meng2019NeuralAutoNavigation}, or learn with a trivial task map~\cite{rana2019learning}. 
By contrast, using our proposed \alg algorithm, we can easily combine gradient-based learning algorithm with arbitrary task or RMP parameterizations.

\textbf{Reinforcement Learning: } In reinforcement learning (RL) applications, one can choose whether to differentiate through the \alg algorithm: one can either choose \alg as part of the policy, or as a component of the environment dynamics. This choice can be made in consideration of the policy parameterization. For example, if a large number of leaf RMPs are parameterized, it could be beneficial to consider \alg as part of the policy and differentiate through it, because otherwise it will result in a high-dimensional action space for RL. On the other hand, if \alg is considered as part of the environment, the output of the parameterized RMPs or parameterized task maps are treated as the action in RL, and there is no need to differentiate through \alg. 
When there is only a single low-dimensional leaf RMP to be learned, it might be convenient to consider \alg as part of the environment so that policy update is faster.\footnote{The time for differentiating through \alg is a constant factor more than the time required for computing \alg, which is still reasonably fast.} Recently \citet{aljalbout2020learning} explored learning collision avoidance RMPs with \flow as a component of the environment. 

\subsection{Learning with Residuals} 

For many robotics tasks, hand-crafted RMPs~\citep{cheng2018rmpflow} can provide a possibly sub-optimal but informative prior solution to the task or some subtasks (e.g. avoiding collision with obstacles, respecting joint limits, etc.). In many cases, making use of these hand-crafted RMPs within \alg policies can benefit learning, as it could provide a reasonable initialization for the learner and, for RL, an initial state visitation distribution with higher rewards. 

\textbf{Residual Acceleration Learning: } Perhaps the most straightforward solution is to learn the residual policy of the \alg policy using universal functional approximators, e.g. neural networks~\cite{johannink2019residual}. As we show in the experiments, this can often provide a significant improvement to the performance compared to randomly initialized policies, especially when the tasks are more challenging. 

\textbf{Residual RMP learning: } Another option is to learn a residual leaf RMP with a universal functional approximator (i.e. the leaf RMP is initialized as the hand-crafted RMP). In this way, the structure of the \alg policy is preserved. As is shown in the experiments, residual RMP learning can perform significantly better than randomly initialized neural network policies, and can sometimes learn faster than the residual acceleration learning approach.

\vspace{-0.5mm}
\section{Experiments}\label{sec:experiments}
\vspace{-0.5mm}
\begin{figure*}
    \centering
    \includegraphics[trim={0px 35px 0px 0px},clip, height=90px]{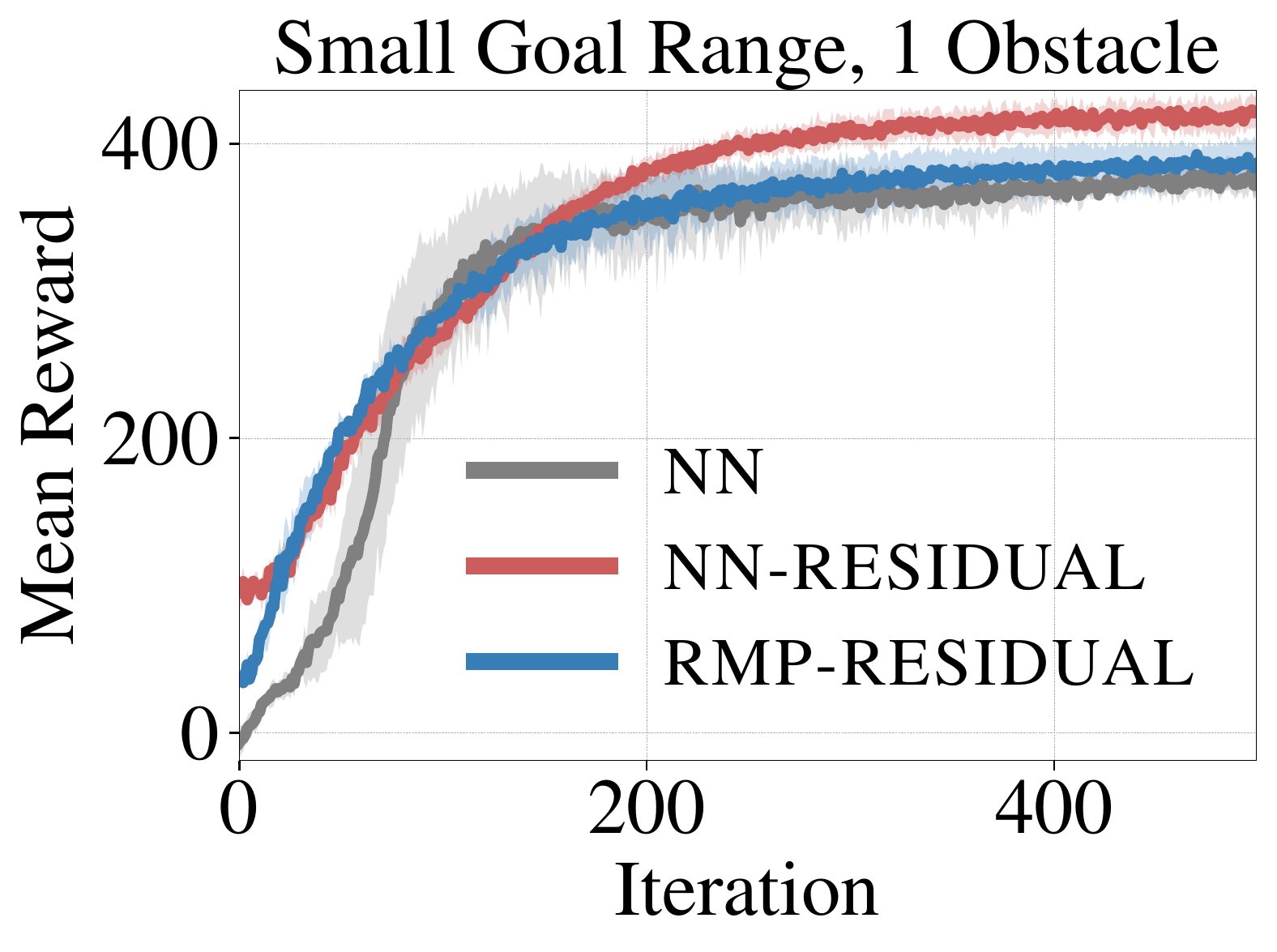}
    \includegraphics[trim={27px 35px 0px 0px},clip, height=90px]{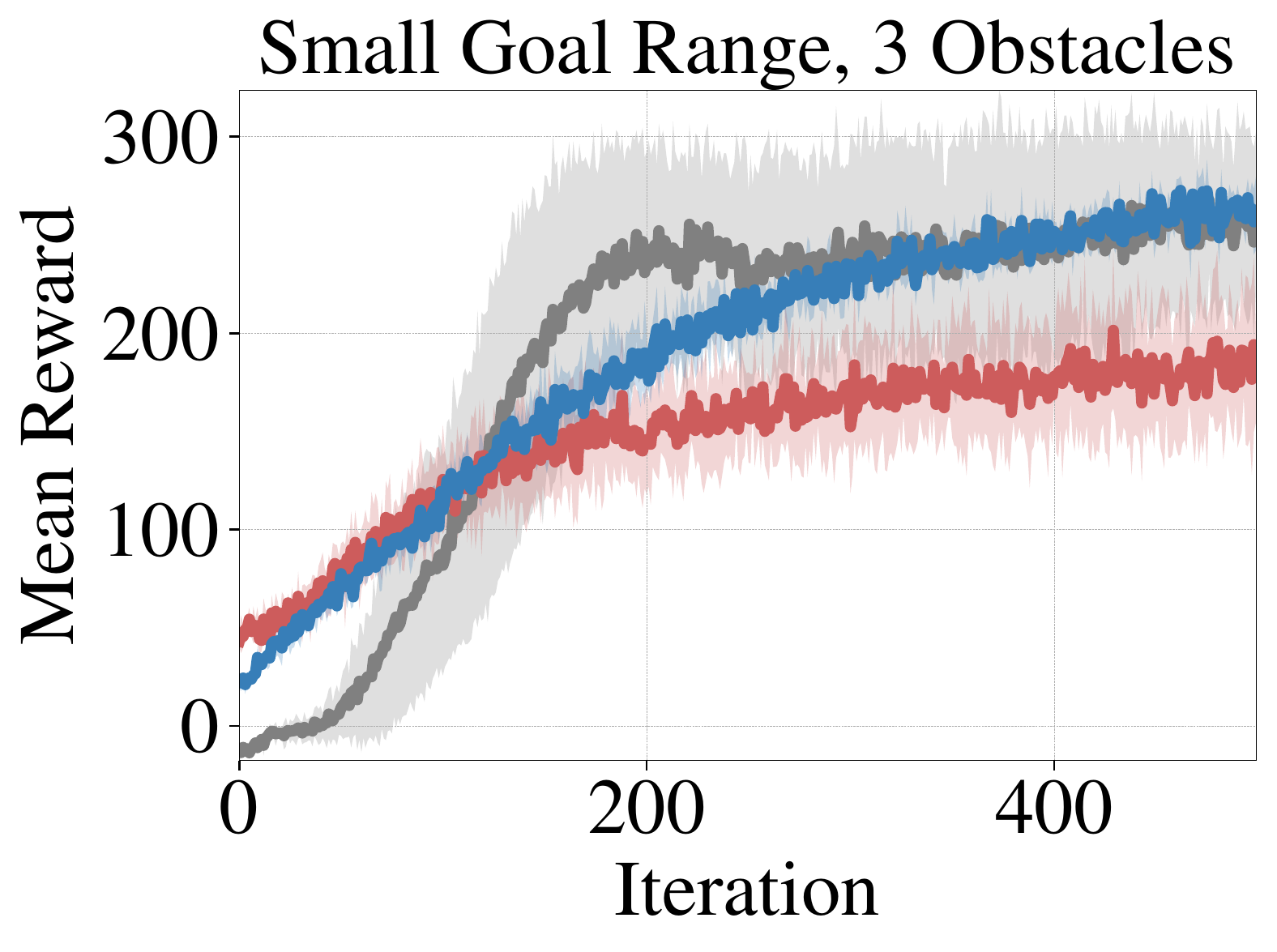}
    \includegraphics[trim={27px 35px 0px 0px},clip, height=90px]{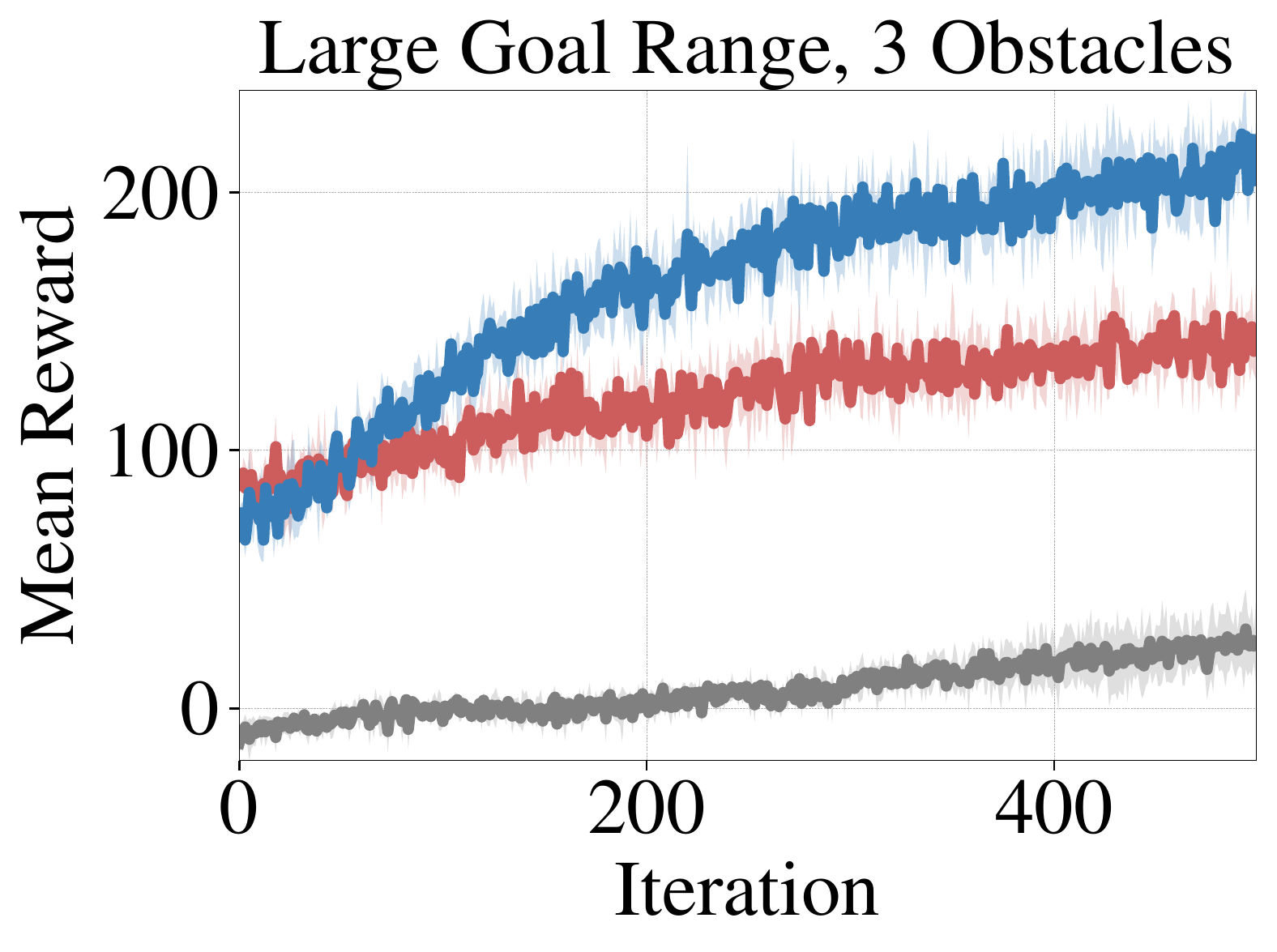}
    \caption{Mean episode reward over training iterations for the three-link robot reaching task. See text for details. }
    \label{fig:reward}
\end{figure*}

\begin{figure*}
    \centering
    \includegraphics[trim={0px 35px 0px 0px},clip, height=93px]{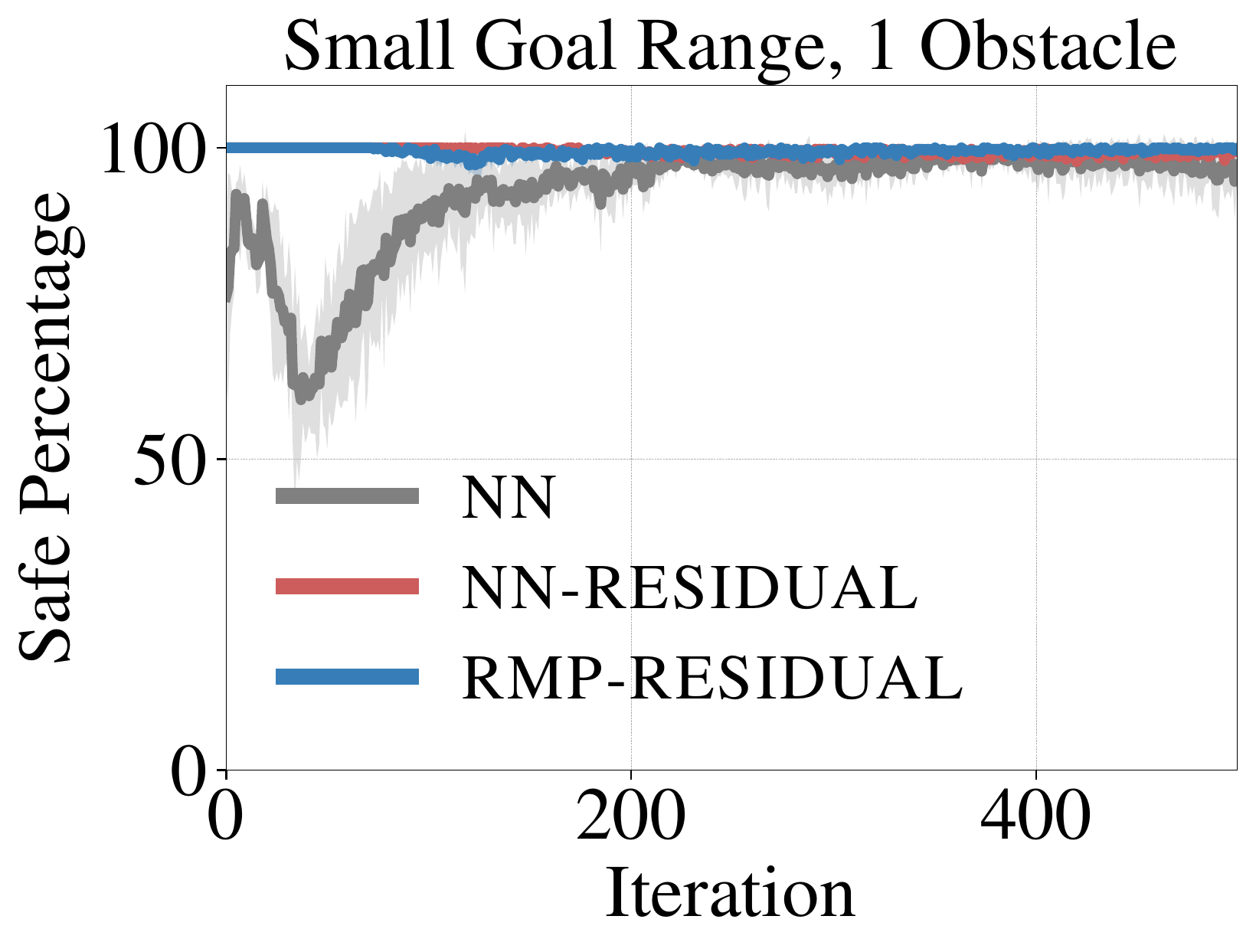}
    \includegraphics[trim={34px 35px 0px 0px},clip, height=93px]{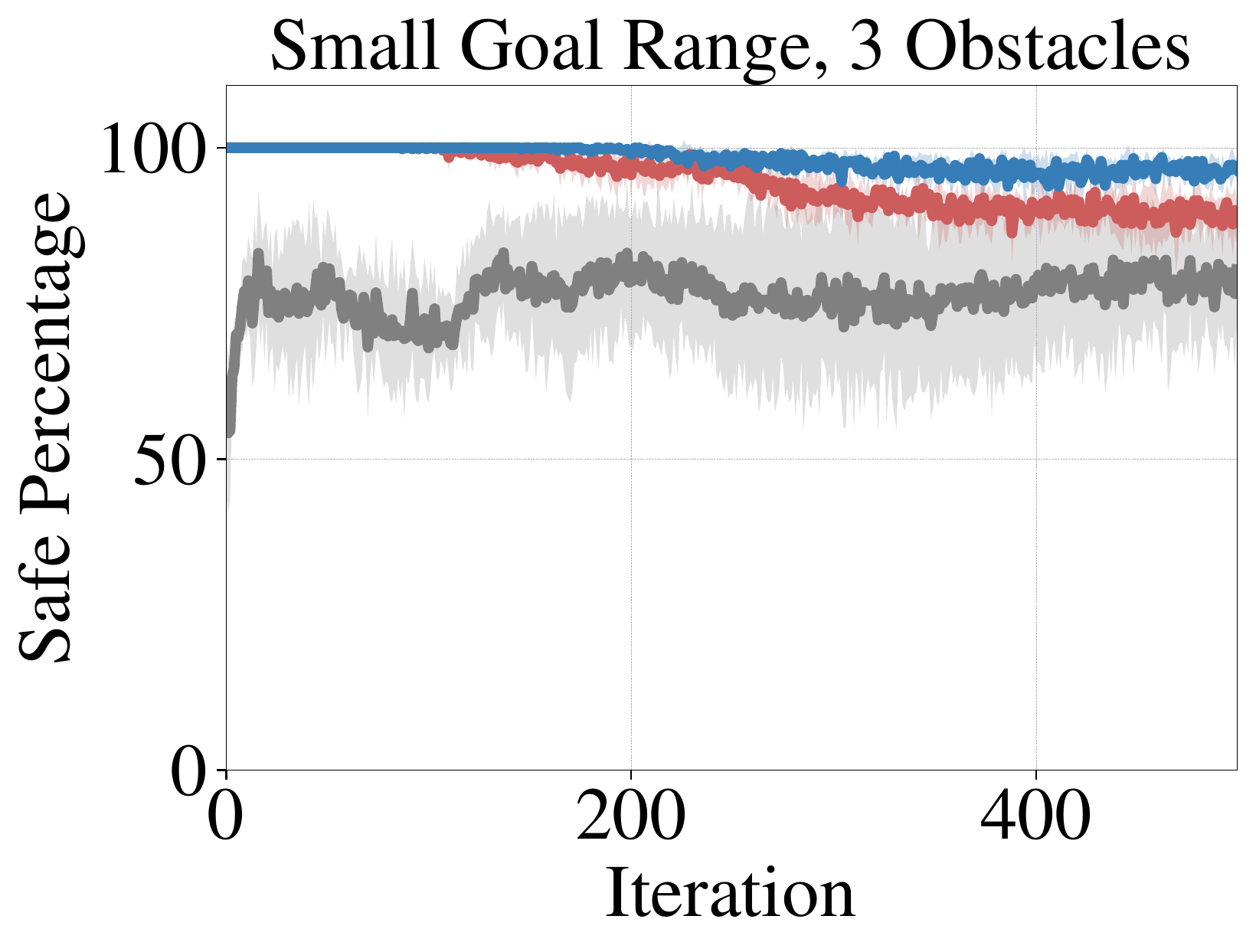}
    \includegraphics[trim={34px 35px 0px 0px},clip, height=93px]{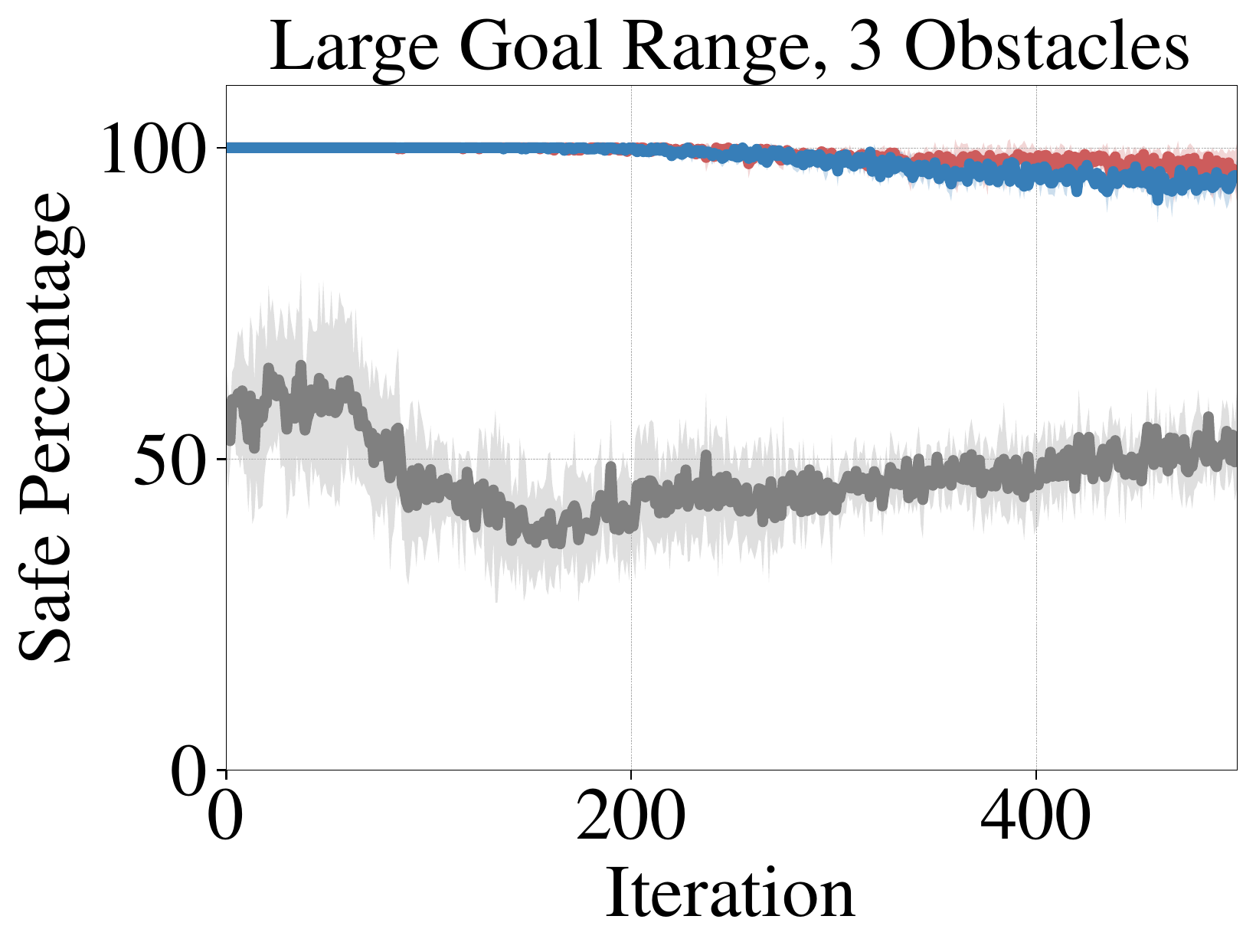}
    \caption{Percentage of safe episodes over training iterations for the Franka robot reaching task. See text for details.}
    \label{fig:safe}
\end{figure*}

\begin{figure*}
    \centering
    \includegraphics[trim={68px 0px 34px 0px},clip,height=90px]{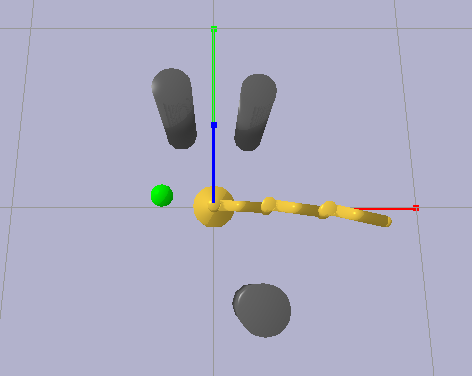}
    \includegraphics[height=90px]{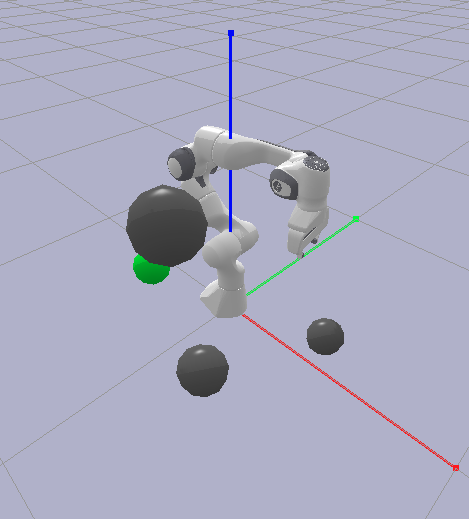}
    \includegraphics[trim={0px 35px 0px 0px},clip,height=90px]{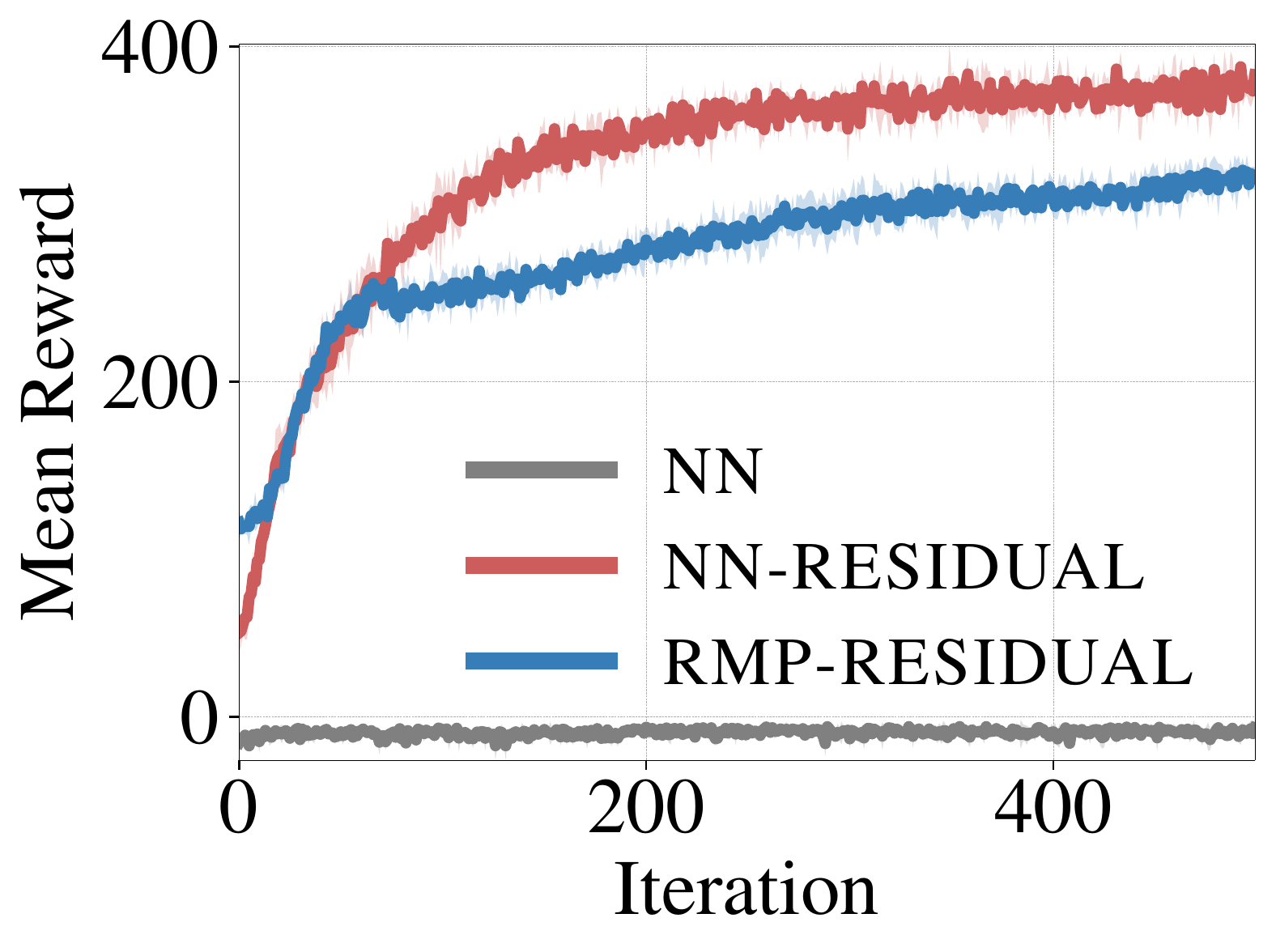}
    \includegraphics[trim={0px 35px 0px 0px},clip,height=90px]{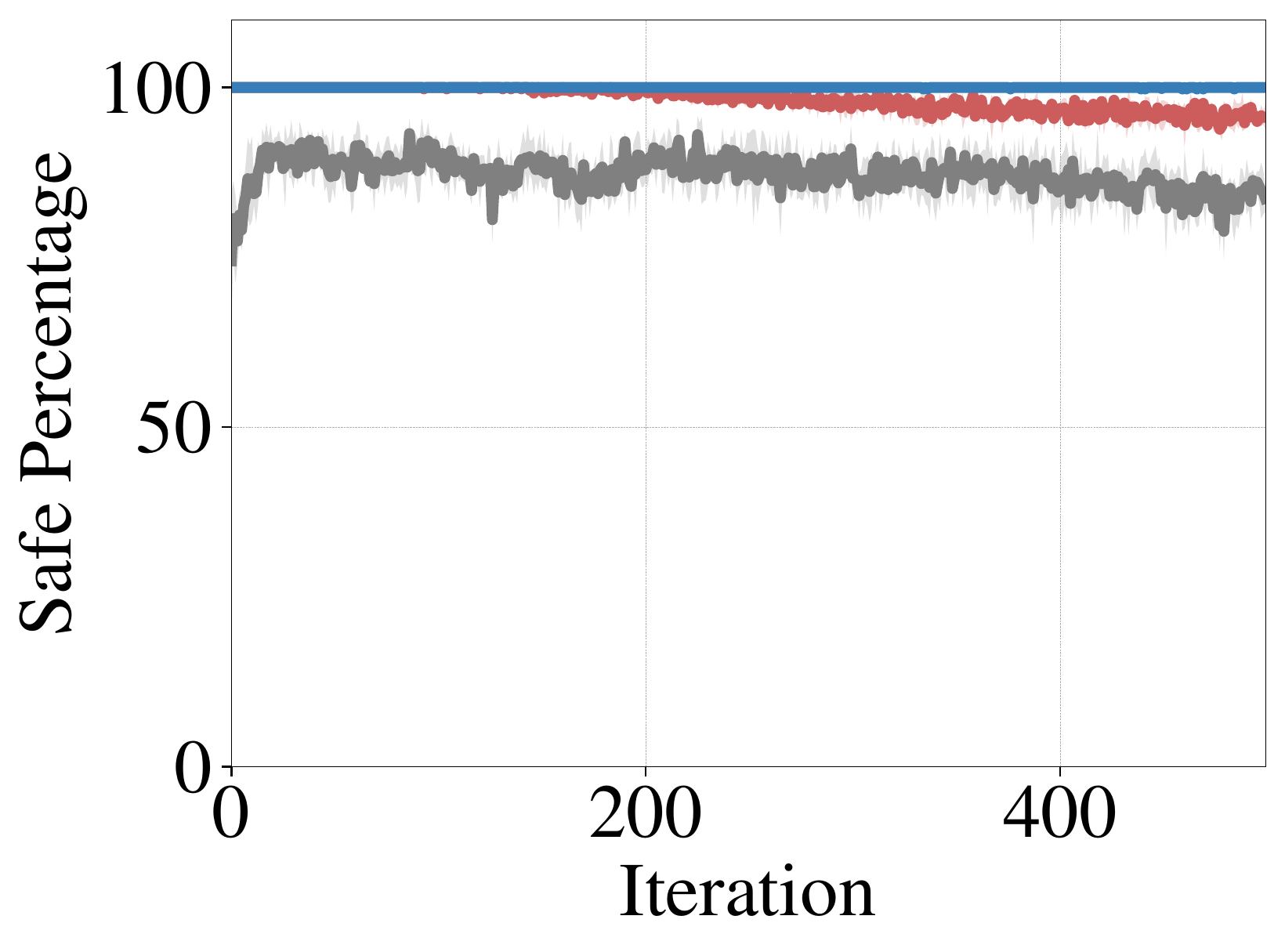}
    \caption{(Left) PyBullet simulation for the  three-link robot reaching task and the Franka robot reachng task. (Right) Mean episode reward and safe episode percentage over training iterations for the Franka reaching task.}
    \label{fig:franka}
\end{figure*}

\subsection{Three-Link Robot Reaching}

We first consider a three-link robot simulated by the PyBullet physics engine~\cite{coumans2017pybullet}. The robot has $3$ links, each of length $0.25$ m. The workspace of the robot is a 2-dimensional disk of radius $0.75$ m. The $z$-coordinates of all links are different so that the links cannot collide into one another. The objective here is for the robot to move the tip of the last link (i.e., the end-effector of the robot) to a randomly generated goal location while avoiding randomly generated cylinders, as shown in~\cref{fig:franka} (left). Acceleration-based control is realized through a low-level PD tracking controller, where the acceleration motion policy $\pi$ generates the desired reference state for the PD controller. The robot does not have joint limits but have joint velocity limit of $1.0$ rad/s for all joints.

\textbf{Environment Setups}
The robot is initialized at a random configuration within a small range $(\pm 0.1 \text{rad})$ around the zero-configuration (all links pointing right) and at a random low joint velocity within $[-0.005, 0.005]$ rad/s. 

The (2-dimensional) center of the base of each obstacle is sampled from an annulus centered at the origin with outer radius $0.9$ m and inner radius $0.4$ m, and the radius is sampled uniformly from $[0.05, 0.1]$ m. The height of the cylinder is fixed at $0.5$ m, which would result in collision if the $x,y$-coordinates of the robot intersects with a cylinder. The consideration behind this obstacle configuration is that their intersection with the workspace is usually non-zero, and they would not result in a configuration where the goal is not achievable. The initial configurations of the environment also ensures a minimum of $0.1$ m between the goal and any obstacles as well as that between the robot and any obstacles (otherwise, the initialization is rejected, and the goal and obstacle(s) are re-sampled). 

We consider three environment setups for the three-link robot with increasing difficulty: 
\begin{itemize}
    \item \texttt{Env 1} (small goal range; 1 obstacle): the goal is sampled from the intersection of an octants (sector with central angle $\pi/2$) at the origin and an annulus with outer radius $0.275$ m and inner radius $0.475$ m. One obstacle is sampled from the procedure described above; 
    \item \texttt{Env 2} (small goal range; 3 obstacles): the goal is sampled from the same region as Env 1. Three obstacles are sampled through the same procedure;
    \item \texttt{Env 3} (large goal range; 3 obstacles): the goal is sampled from the intersection of the left half-disk and an annulus with outer radius $0.125$ m and inner radius $0.625$ m. 
\end{itemize}
\texttt{Env 2} is more difficult than \texttt{Env 1} as there are more obstacles. \texttt{Env 3} is the most complicated scenario as it has a larger range of random goals, which is known to be challenging for RL algorithms~\cite{aumjaud2020reinforcement}. Moreover, although the goals here are generally closer than the previous 2 environment setups, this also increase the frequency of the scenarios where the obstacles are directly obstructing the way to the goal, requiring a more sophisticated policy.

Inspired by~\cite{kumar2020joint}, we define the reward function as,
$    r = \exp\left(\frac{\|\x-\g\|_2^2}{2\sigma^2}\right) + \sum_{i=1}^{N}\max\left(0, 1-\frac{d_i}{\delta}\right) - \lambda \|\tau\|^2$, 
where $\x$ is the position of the end-effector of the robot, $\g$ is the goal position, $d_i$ is the distance between the robot and the $i$th obstacle, and $\tau$ is the torque applied to the robot by the low-level PD controller. The scalars $\sigma$, $\delta$, and $\lambda$ are the characteristic length scale of the goal reward, characteristic length scale of the obstacle cost, and the multiplier for the actuation cost. We choose $\sigma=0.1$, $\delta=0.05$, and $\lambda=1\times 10^{-5}$. Further, we clip the reward if it is smaller than $-5$ so that the reward is in the range of $[-5, 1]$ for each step.

The horizon of each episode is $600$ steps, giving the episode reward a range of $[-3000, 600]$. For each step, the acceleration command is applied to the low-level PD-controller, and the simulation proceeds for $0.0125$s (simulation time). Therefore, each episode is $7.5$s (simulation time) of policy rollout. The episode can end early if the robot collides with an obstacle. 

\textbf{Policies}
We compare the results for learning with the following 3 types of policies, all implemented in TensorFlow~\cite{tensorflow2015-whitepaper}. 
    
    {NN}: A randomly initialized 3-layer neural network policy with $\mathrm{relu}$ activation function and hidden layers of sizes 256 and 128, respectively. The input to the neural network policy consists of $[\sin(\q),\,\cos(\q),\,\qd,\,\g-\x,\,\{\vb_i\}_i,\,\{\ob_i\}_i],$
    where $\vb_i$ is the vector pointing from the $i$-th obstacle and its closest point on the robot, and $\ob_i$ denotes the center position and radius of the $i$-th obstacle. The output of the policy is the 3-dimensional joint acceleration. 
    
    {NN-RESIDUAL}: A residual neural network policy to a hand-designed \alg policy. The input and neural network architecture here is the same as the randomly-initialized case above. However, the joint acceleration now is the sum of the output of the residual policy and the hand-designed \alg policy. The hand-designed \alg policy consists of a joint damping RMP, a joint velocity limit RMP, collision avoidance RMPs, and a goal attractor RMP~\cite{cheng2018rmpflow}. 

    {RMP-RESIDUAL}: A residual RMP policy on the 2-dimensional end-effector space, which is the residual to a hand-designed goal attractor RMP (same as the one used for NN-RESIDUAL). 
    The residual RMP policy is parameterized as a 3-layer neural network with hidden layer sizes $128$ and $64$. We use $\mathrm{elu}$ activation function for the neural network. The input to the residual RMP policy consists of $[\x,\xd,\g, \{\ob_i\}_i]$, which are the end effector position, velocity, and the information of goal and obstacle(s). The output of the neural network is (the concatenation of) a matrix $\Ab_r\in\R^{2\times 2}$ and a residual acceleration vector $\ab_r\in\R^2$. Let $(\M_a, \ab_a)$ be the output for the hand-designed attractor, the output of the residual RMP policy is then,
    $        \M = (\Ab_r + \texttt{chol}(\M_a))(\Ab_r + \texttt{chol}(\M_a))^\t$ and 
    $\ab = \ab_r + \ab_a$, 
    where $\texttt{chol}(\cdot)$ is the lower-triangular Cholesky decomposition of the matrix. This parameterization ensures that the importance weight $\M$ is always positive-semidefinite, and the output is close to the hand-designed RMP when the neural network is initialized with weights close to zero. The output of the residual RMP policy, is combined with other hand-designed RMPs (the same set of RMPs with NN-RESIDUAL) with \alg to produce the joint acceleration. Since we are learning a low-dimensional RMP, as is discussed in~\cref{sec:learning}, it is more convenient to consider \alg as part of the environment so that the policy update is faster. 

\textbf{Training Details}
We train the three policies under the three environment setups by Proximal Policy Optimization (PPO)~\cite{schulman2017proximal}, implemented by RLlib~\cite{liang2018rllib},  for $500$ training iterations. For each iteration, we collect a batch of $67312$ interactions with the environment (112 episodes if there is not collision). We use a learning rate of $5\times 10^{-5}$, PPO clip parameter of $0.2$. To compute the policy gradient, we use Generalized Advantage Estimation (GAE)~\cite{schulman2015high} with $\lambda=0.99$. The value function is parameterized by a neural network with 2 hidden layers (of sizes 256 and 128, respectively) and $\tanh$ activation function. 

\textbf{Results}
The mean episode reward and percentage of safe episodes over training iterations for the three environment setups are shown in~\cref{fig:reward} and~\cref{fig:safe}, respectively. The curve and the shaded region show the average, and the region within 1 standard deviation from the mean over 4 random seeds. 

For \texttt{env 1} with small goal range and 1 obstacle, all three policies manage to achieve high reward of around 400 (meaning that, on average, the robot stays very close to the goal for at least 400 steps out of the 600 steps) with a similar number of iterations. The random-initialized neural network (NN) conducts a large number of unsafe exploration, especially during the first 200 iterations, as shown in~\cref{fig:safe} (left). The other two policies manage to learn a good policy without many collisions as the red and blue curves stay close to 100 throughout training. 

For \texttt{env 2} with 3 obstacles, the randomly-initialized neural network policy (NN) learns slightly faster than the other two polices, though it has a higher variance (shown by the large gray shaded region in~\cref{fig:reward} (middle)). Notably, although the reward seems high, the resulting policies are not safe, as shown in~\cref{fig:safe} (middle). This reflects the difficulty of reward design: as high reward policies do not necessarily have desirable behaviors. On the contrary, the residual RMP policy (RMP-RESIDUAL) achieves similar reward but is able to keep the number of collisions low. The residual neural network policy (NN-RESIDUAL) improves rather slowly, possibly due to the lack of knowledge about the internal structure of~\alg. 

For \texttt{env 3} with large goal range and 3 obstacles, the neural netowrk policy (NN) struggles to achieve reasonable performance under 500 training iterations, and the collision rate remains high. The performance improvement for the reisual neural network policy (NN-RESIDUAL) is also slow, with a slope similar to the neural network policy, though it starts with a higher rewards. The residual RMP policy manages to improve the performance by more than one-fold. 

\textbf{Remark: }
One may notice that the initial reward for the residual policies is different for each experiment setup. For example, the initial rewards for \texttt{env 3} is significantly higher than \texttt{env 2}. This is because we used the \emph{same} hand-designed RMP policy for all three setups, and in \texttt{env 3}, the goals are generally initialized closer to the robot than the other two steps (as discussed, this does not make \texttt{env 3} easier however).

\subsection{Franka Robot Reaching}

We additionally train the three aforementioned policies on a simulated 7-degree-of-freedom Franka manipulator on PyBullet. The environment setup is similar to the three-link robot reaching task, although the initial configuration is a centered position, as shown in~\ref{fig:franka}. We randomly sample 3 ball obstacles, where the center is sampled a half-torus of major radius 0.5 m, minor radius 0.3 m, and height 0.5 m; and the radius is uniformly sampled from $[0.05, 0.1]$ m. The goal is sampled from the same half-torus with a ball centered at the initial end-effector position and radius $0.5$ m excluded, meaning that the distance between the goal and initial end-effector position should be at least $0.5$ m. The Franka reaching task is more challenging than the three-link robot reaching task as the states are of higher-dimension; however, it is easier in the sense that it has higher degrees of freedom to avoid collision with obstacles. 

\textbf{Results}
The mean episode reward and safe percentage of the three polices for the Franka reaching task is shown in~\cref{fig:franka} (right 2). Again, the neural network policy (NN) struggles to learn a good policy and avoid collision with obstacles. The residual neural network policy (NN-RESIDUAL) achieves slightly higher reward than the residual RMP policy (RMP-RESIDUAL), possibly because it has higher degree-of-freedoms to control the robot and the learning of the residual RMP policy (RMP-RESIDUAL) has not fully converged. Notably, the residual RMP policy (RMP-RESIDUAL) maintain zero collisions throughout the training process. The visualization of the learned policies for both experiments can be found at \url{https://youtu.be/dliQ-jsYhgI}.


\bibliographystyle{abbrvnat}
\bibliography{refs}

\begin{thebibliography}{38}
\providecommand{\natexlab}[1]{#1}
\providecommand{\url}[1]{\texttt{#1}}
\expandafter\ifx\csname urlstyle\endcsname\relax
  \providecommand{\doi}[1]{doi: #1}\else
  \providecommand{\doi}{doi: \begingroup \urlstyle{rm}\Url}\fi

\bibitem[Abadi et~al.(2015)Abadi, Agarwal, Barham, Brevdo, Chen, Citro,
  Corrado, Davis, Dean, Devin, Ghemawat, Goodfellow, Harp, Irving, Isard, Jia,
  Jozefowicz, Kaiser, Kudlur, Levenberg, Man\'{e}, Monga, Moore, Murray, Olah,
  Schuster, Shlens, Steiner, Sutskever, Talwar, Tucker, Vanhoucke, Vasudevan,
  Vi\'{e}gas, Vinyals, Warden, Wattenberg, Wicke, Yu, and
  Zheng]{tensorflow2015-whitepaper}
M.~Abadi, A.~Agarwal, P.~Barham, E.~Brevdo, Z.~Chen, C.~Citro, G.~S. Corrado,
  A.~Davis, J.~Dean, M.~Devin, S.~Ghemawat, I.~Goodfellow, A.~Harp, G.~Irving,
  M.~Isard, Y.~Jia, R.~Jozefowicz, L.~Kaiser, M.~Kudlur, J.~Levenberg,
  D.~Man\'{e}, R.~Monga, S.~Moore, D.~Murray, C.~Olah, M.~Schuster, J.~Shlens,
  B.~Steiner, I.~Sutskever, K.~Talwar, P.~Tucker, V.~Vanhoucke, V.~Vasudevan,
  F.~Vi\'{e}gas, O.~Vinyals, P.~Warden, M.~Wattenberg, M.~Wicke, Y.~Yu, and
  X.~Zheng.
\newblock {TensorFlow}: Large-scale machine learning on heterogeneous systems,
  2015.
\newblock URL \url{http://tensorflow.org/}.
\newblock Software available from tensorflow.org.

\bibitem[Aljalbout et~al.(2020)Aljalbout, Chen, Ritt, Ulmer, and
  Haddadin]{aljalbout2020learning}
E.~Aljalbout, J.~Chen, K.~Ritt, M.~Ulmer, and S.~Haddadin.
\newblock Learning vision-based reactive policies for obstacle avoidance.
\newblock \emph{arXiv preprint arXiv:2010.16298}, 2020.

\bibitem[Aumjaud et~al.(2020)Aumjaud, McAuliffe, Rodr{\'\i}guez-Lera, and
  Cardiff]{aumjaud2020reinforcement}
P.~Aumjaud, D.~McAuliffe, F.~J. Rodr{\'\i}guez-Lera, and P.~Cardiff.
\newblock Reinforcement learning experiments and benchmark for solving robotic
  reaching tasks.
\newblock In \emph{Workshop of Physical Agents}, pages 318--331. Springer,
  2020.

\bibitem[Bullo and Lewis(2005)]{bullo2005geometric}
F.~Bullo and A.~D. Lewis.
\newblock \emph{Geometric control of mechanical systems: modeling, analysis,
  and design for simple mechanical control systems}.
\newblock Springer New York, 2005.

\bibitem[Cheng(2020)]{cheng2020efficient}
C.~A. Cheng.
\newblock \emph{Efficient and principled robot learning: theory and
  algorithms}.
\newblock PhD thesis, Georgia Institute of Technology, 2020.

\bibitem[Cheng et~al.(2018)Cheng, Mukadam, Issac, Birchfield, Fox, Boots, and
  Ratliff]{cheng2018rmpflow}
C.-A. Cheng, M.~Mukadam, J.~Issac, S.~Birchfield, D.~Fox, B.~Boots, and
  N.~Ratliff.
\newblock {RMP}flow: A computational graph for automatic motion policy
  generation.
\newblock \emph{Proceedings of the 13th Annual Workshop on the Algorithmic
  Foundations of Robotics (WAFR)}, 2018.

\bibitem[Cheng et~al.(2020)Cheng, Mukadam, Issac, Birchfield, Fox, Boots, and
  Ratliff]{cheng2020rmpflow}
C.-A. Cheng, M.~Mukadam, J.~Issac, S.~Birchfield, D.~Fox, B.~Boots, and
  N.~Ratliff.
\newblock {RMP}flow: A geometric framework for generation of multi-task motion
  policies.
\newblock \emph{arXiv preprint arXiv:2007.14256}, 2020.

\bibitem[Choi et~al.(2020)Choi, Casta{\~n}eda, Tomlin, and
  Sreenath]{choi2020reinforcement}
J.~Choi, F.~Casta{\~n}eda, C.~J. Tomlin, and K.~Sreenath.
\newblock Reinforcement learning for safety-critical control under model
  uncertainty, using control {L}yapunov functions and control barrier
  functions.
\newblock \emph{arXiv preprint arXiv:2004.07584}, 2020.

\bibitem[Chow et~al.(2019)Chow, Nachum, Faust, Duenez-Guzman, and
  Ghavamzadeh]{chow2019lyapunov}
Y.~Chow, O.~Nachum, A.~Faust, E.~Duenez-Guzman, and M.~Ghavamzadeh.
\newblock Lyapunov-based safe policy optimization for continuous control.
\newblock \emph{arXiv preprint arXiv:1901.10031}, 2019.

\bibitem[Christianson(1992)]{christianson1992automatic}
B.~Christianson.
\newblock Automatic hessians by reverse accumulation.
\newblock \emph{IMA Journal of Numerical Analysis}, 12\penalty0 (2):\penalty0
  135--150, 1992.

\bibitem[Coumans and Bai(2017)]{coumans2017pybullet}
E.~Coumans and Y.~Bai.
\newblock Py{B}ullet, a python module for physics simulation in robotics, games
  and machine learning, 2017.

\bibitem[Dalal et~al.(2018)Dalal, Dvijotham, Vecerik, Hester, Paduraru, and
  Tassa]{dalal2018safe}
G.~Dalal, K.~Dvijotham, M.~Vecerik, T.~Hester, C.~Paduraru, and Y.~Tassa.
\newblock Safe exploration in continuous action spaces.
\newblock \emph{arXiv preprint arXiv:1801.08757}, 2018.

\bibitem[Griewank and Walther(2008)]{griewank2008evaluating}
A.~Griewank and A.~Walther.
\newblock \emph{Evaluating derivatives: principles and techniques of
  algorithmic differentiation}.
\newblock SIAM, 2008.

\bibitem[Johannink et~al.(2019)Johannink, Bahl, Nair, Luo, Kumar, Loskyll,
  Ojea, Solowjow, and Levine]{johannink2019residual}
T.~Johannink, S.~Bahl, A.~Nair, J.~Luo, A.~Kumar, M.~Loskyll, J.~A. Ojea,
  E.~Solowjow, and S.~Levine.
\newblock Residual reinforcement learning for robot control.
\newblock In \emph{2019 International Conference on Robotics and Automation
  (ICRA)}, pages 6023--6029. IEEE, 2019.

\bibitem[Kappler et~al.(2018)Kappler, Meier, Issac, Mainprice,
  Garcia~Cifuentes, W{\"u}thrich, Berenz, Schaal, Ratliff, and
  Bohg]{2017_rss_system}
D.~Kappler, F.~Meier, J.~Issac, J.~Mainprice, C.~Garcia~Cifuentes,
  M.~W{\"u}thrich, V.~Berenz, S.~Schaal, N.~Ratliff, and J.~Bohg.
\newblock Real-time perception meets reactive motion generation.
\newblock \emph{IEEE Robotics and Automation Letters}, 3\penalty0 (3):\penalty0
  1864--1871, 2018.
\newblock URL \url{https://arxiv.org/abs/1703.03512}.

\bibitem[Khalil and Grizzle(2002)]{khalil2002nonlinear}
H.~K. Khalil and J.~W. Grizzle.
\newblock \emph{Nonlinear systems}, volume~3.
\newblock Prentice hall Upper Saddle River, NJ, 2002.

\bibitem[Kumar et~al.(2020)Kumar, Hoeller, Sundaralingam, Tremblay, and
  Birchfield]{kumar2020joint}
V.~Kumar, D.~Hoeller, B.~Sundaralingam, J.~Tremblay, and S.~Birchfield.
\newblock Joint space control via deep reinforcement learning.
\newblock \emph{arXiv preprint arXiv:2011.06332}, 2020.

\bibitem[Li et~al.(2019{\natexlab{a}})Li, Cheng, Boots, and
  Egerstedt]{li2019stable}
A.~Li, C.-A. Cheng, B.~Boots, and M.~Egerstedt.
\newblock Stable, concurrent controller composition for multi-objective robotic
  tasks.
\newblock In \emph{2019 IEEE 58th Conference on Decision and Control (CDC)},
  pages 1144--1151. IEEE, 2019{\natexlab{a}}.

\bibitem[Li et~al.(2019{\natexlab{b}})Li, Mukadam, Egerstedt, and
  Boots]{li2019MultiAgentRMPsArXiv}
A.~Li, M.~Mukadam, M.~Egerstedt, and B.~Boots.
\newblock Multi-objective policy generation for multi-robot systems using
  {R}iemannian motion policies.
\newblock In \emph{International Symposium on Robotics Research},
  2019{\natexlab{b}}.

\bibitem[Liang et~al.(2018)Liang, Liaw, Nishihara, Moritz, Fox, Goldberg,
  Gonzalez, Jordan, and Stoica]{liang2018rllib}
E.~Liang, R.~Liaw, R.~Nishihara, P.~Moritz, R.~Fox, K.~Goldberg, J.~Gonzalez,
  M.~Jordan, and I.~Stoica.
\newblock R{L}lib: Abstractions for distributed reinforcement learning.
\newblock In \emph{International Conference on Machine Learning}, pages
  3053--3062. PMLR, 2018.

\bibitem[Meng et~al.(2019)Meng, Ratliff, Xiang, and
  Fox]{meng2019NeuralAutoNavigation}
X.~Meng, N.~Ratliff, Y.~Xiang, and D.~Fox.
\newblock Neural autonomous navigation with {R}iemannian motion policy.
\newblock In \emph{2019 International Conference on Robotics and Automation
  (ICRA)}, pages 8860--8866. IEEE, 2019.

\bibitem[Mukadam et~al.(2019)Mukadam, Cheng, Fox, Boots, and
  Ratliff]{mukadam2019riemannian}
M.~Mukadam, C.-A. Cheng, D.~Fox, B.~Boots, and N.~Ratliff.
\newblock Riemannian motion policy fusion through learnable {L}yapunov function
  reshaping.
\newblock In \emph{Conference on Robot Learning}, pages 204--219, 2019.

\bibitem[Nakanishi et~al.(2008)Nakanishi, Cory, Mistry, Peters, and
  Schaal]{nakanishi2008operational}
J.~Nakanishi, R.~Cory, M.~Mistry, J.~Peters, and S.~Schaal.
\newblock Operational space control: A theoretical and empirical comparison.
\newblock \emph{The International Journal of Robotics Research}, 27\penalty0
  (6):\penalty0 737--757, 2008.

\bibitem[Paszke et~al.(2017)Paszke, Gross, Chintala, Chanan, Yang, DeVito, Lin,
  Desmaison, Antiga, and Lerer]{paszke2017automatic}
A.~Paszke, S.~Gross, S.~Chintala, G.~Chanan, E.~Yang, Z.~DeVito, Z.~Lin,
  A.~Desmaison, L.~Antiga, and A.~Lerer.
\newblock Automatic differentiation in {PyTorch}.
\newblock In \emph{NIPS Autodiff Workshop}, 2017.

\bibitem[Paxton et~al.(2019)Paxton, Ratliff, Eppner, and Fox]{paxton2019RLDS}
C.~Paxton, N.~Ratliff, C.~Eppner, and D.~Fox.
\newblock Representing robot task plans as robust logical-dynamical systems.
\newblock In \emph{2019 IEEE/RSJ International Conference on Intelligent Robots
  and Systems (IROS)}, 2019.

\bibitem[Pomerleau(1988)]{pomerleau1988alvinn}
D.~A. Pomerleau.
\newblock {ALVINN}: an autonomous land vehicle in a neural network.
\newblock In \emph{Proceedings of the 1st International Conference on Neural
  Information Processing Systems}, pages 305--313, 1988.

\bibitem[Rana et~al.(2019)Rana, Li, Ravichandar, Mukadam, Chernova, Fox, Boots,
  and Ratliff]{rana2019learning}
M.~A. Rana, A.~Li, H.~Ravichandar, M.~Mukadam, S.~Chernova, D.~Fox, B.~Boots,
  and N.~Ratliff.
\newblock Learning reactive motion policies in multiple task spaces from human
  demonstrations.
\newblock In \emph{Conference on Robot Learning}, 2019.

\bibitem[Rana et~al.(2020{\natexlab{a}})Rana, Li, Fox, Boots, Ramos, and
  Ratliff]{rana2020euclideanizing}
M.~A. Rana, A.~Li, D.~Fox, B.~Boots, F.~Ramos, and N.~Ratliff.
\newblock Euclideanizing flows: Diffeomorphic reduction for learning stable
  dynamical systems.
\newblock In \emph{Learning for Dynamics and Control}, pages 630--639. PMLR,
  2020{\natexlab{a}}.

\bibitem[Rana et~al.(2020{\natexlab{b}})Rana, Li, Fox, Chernova, Boots, and
  Ratliff]{rana2020towards}
M.~A. Rana, A.~Li, D.~Fox, S.~Chernova, B.~Boots, and N.~Ratliff.
\newblock Towards coordinated robot motions: End-to-end learning of motion
  policies on transform trees.
\newblock \emph{arXiv preprint arXiv:2012.13457}, 2020{\natexlab{b}}.

\bibitem[Ratliff et~al.(2018)Ratliff, Issac, Kappler, Birchfield, and
  Fox]{ratliff2018riemannian}
N.~D. Ratliff, J.~Issac, D.~Kappler, S.~Birchfield, and D.~Fox.
\newblock Riemannian motion policies.
\newblock \emph{arXiv preprint arXiv:1801.02854}, 2018.

\bibitem[Ross and Bagnell(2014)]{ross2014reinforcement}
S.~Ross and J.~A. Bagnell.
\newblock Reinforcement and imitation learning via interactive no-regret
  learning.
\newblock \emph{arXiv preprint arXiv:1406.5979}, 2014.

\bibitem[Ross et~al.(2011)Ross, Gordon, and Bagnell]{ross2011reduction}
S.~Ross, G.~Gordon, and D.~Bagnell.
\newblock A reduction of imitation learning and structured prediction to
  no-regret online learning.
\newblock In \emph{Proceedings of the fourteenth international conference on
  artificial intelligence and statistics}, pages 627--635. JMLR Workshop and
  Conference Proceedings, 2011.

\bibitem[Schulman et~al.(2015)Schulman, Moritz, Levine, Jordan, and
  Abbeel]{schulman2015high}
J.~Schulman, P.~Moritz, S.~Levine, M.~Jordan, and P.~Abbeel.
\newblock High-dimensional continuous control using generalized advantage
  estimation.
\newblock \emph{arXiv preprint arXiv:1506.02438}, 2015.

\bibitem[Schulman et~al.(2017)Schulman, Wolski, Dhariwal, Radford, and
  Klimov]{schulman2017proximal}
J.~Schulman, F.~Wolski, P.~Dhariwal, A.~Radford, and O.~Klimov.
\newblock Proximal policy optimization algorithms.
\newblock \emph{arXiv preprint arXiv:1707.06347}, 2017.

\bibitem[Siciliano et~al.(2010)Siciliano, Sciavicco, Villani, and
  Oriolo]{siciliano2010robotics}
B.~Siciliano, L.~Sciavicco, L.~Villani, and G.~Oriolo.
\newblock \emph{Robotics: modelling, planning and control}.
\newblock Springer Science \& Business Media, 2010.

\bibitem[Sutanto et~al.(2019)Sutanto, Ratliff, Sundaralingam, Chebotar, Su,
  Handa, and Fox]{sutanto2019TactileServoing}
G.~Sutanto, N.~Ratliff, B.~Sundaralingam, Y.~Chebotar, Z.~Su, A.~Handa, and
  D.~Fox.
\newblock Learning latent space dynamics for tactile servoing.
\newblock In \emph{2019 International Conference on Robotics and Automation
  (ICRA)}, 2019.

\bibitem[Urain et~al.(2020)Urain, Ginesi, Tateo, and
  Peters]{urain2020imitationflow}
J.~Urain, M.~Ginesi, D.~Tateo, and J.~Peters.
\newblock Imitationflow: Learning deep stable stochastic dynamic systems by
  normalizing flows.
\newblock \emph{arXiv preprint arXiv:2010.13129}, 2020.

\bibitem[Wingo et~al.(2020)Wingo, Cheng, Murtaza, Zafar, and
  Hutchinson]{wingo2020extending}
B.~Wingo, C.-A. Cheng, M.~Murtaza, M.~Zafar, and S.~Hutchinson.
\newblock Extending {R}iemmanian motion policies to a class of underactuated
  wheeled-inverted-pendulum robots.
\newblock In \emph{2020 IEEE International Conference on Robotics and
  Automation (ICRA)}, pages 3967--3973. IEEE, 2020.

\end{thebibliography}

\newpage

\clearpage
\appendices
\section{The \flow Algorithm}\label{sec:rmpflow_alg}

Here we describe the message passing steps of \flow~\citep{cheng2018rmpflow}. 
As is noted in~\cite{cheng2020efficient},  this message passing routine effectively uses the duality of \eqref{eq:rmpflow_objective} and the sparsity in the task map to efficiently compute $\ab_\rtt$.

The \flow algorithm (\cref{alg:rmpflow}) is based on two components:
\begin{enumerate*}[label=\emph{\arabic*})]
    \item the \tree: a directed tree encoding the structure of the task map; and 
    \item the \algebra: a set of operations to propagate information across the \tree
\end{enumerate*}.

In the \tree, a node $\utt$ stores the state $(\x_\utt, \xd_\utt)$ on a manifold $\MM_\utt$ and the associated RMP $(\ab_\utt, \M_\utt)$. We define the \emph{natural form} of an RMP as $[\fb_\utt, \M_\utt]$, where $\fb_\utt =\Mb_\utt\ab_\utt$ is the force. An edge $\ett$ represents to a smooth map $\psi_{\ett}$ from the manifold of a parent node manifold to its child node manifold. The root node of the \tree (denoted as $\rtt$) and the leaf nodes correspond to the configuration space $\CC$ and the subtask spaces $\{\TT_k\}$ on which the subtask RMPs are hosted, respectively. 

The \algebra comprises of three operators, which are for propagating information on the \tree.
For illustration, we consider the node $\utt$ on manifold $\MM$ with coordinates $\x$ and its $M$ child nodes ($\vtt_m$ with coordinates $\y_m$ for $m=1,\dots,M$) on the \tree. 
\begin{enumerate}[label=(\textit{\roman*})] 
    \item\pushforward propagates the state of a node $(\x,\xd)$ in the \tree to update the states of its $M$ child nodes $\{(\y_m,\yd_m)\}_{m=1}^M$. The state of its $m$th child node is computed as $(\y_m, \yd_m) = (\psi_{\vtt_m;\utt}(\x) , \J_{\vtt_m;\utt}(\x)\,\xd )$,
    where $\psi_m$ is the smooth map of the edge connecting the two nodes and $\J_{\vtt_m;\utt} = \partial_\x \psi_{\vtt_m;\utt}$ is the Jacobian matrix.

    \item \pullback propagates the RMPs of the child nodes in the natural form, $\{[\f_{\vtt_m}, \M_{\vtt_m}]\}_{m=1}^M$, to the parent node as $[\f_\utt, \M_\utt]$:
    \begin{equation}\label{eq:pullback} 
    \begin{split}
        \f_\utt &= \sum_{m=1}^M \J_{\vtt_m;\utt}^\t (\f_{\vtt_m} - \M_{\vtt_m} \Jd_{\vtt_m;\utt}\xd),\\
        \M_\utt &= \sum_{m=1}^M \J_{\vtt_m;\utt}^\t \M_{\vtt_m} \J_{\vtt_m;\utt}.
    \end{split}
    \end{equation}
    The natural form of RMPs are used here since they more efficient to combine.

    \item \resolve maps an RMP from its natural form $[\f_\utt, \M_\utt]$ to its canonical form $(\ab_\utt, \M_\utt)$ by $\ab_\utt = \M_\utt^{\dagger}\,\f_\utt$, where $\dagger$ denotes Moore-Penrose inverse. 

\end{enumerate}
\flow in \cref{alg:rmpflow} computes the policy $\pi(\q(t),\qd(t))=\ab_{\rtt}$ on the configuration space $\CC$ through the following procedure.
Given the state  $(\q(t),\qd(t))$ of the configuration space $\CC$ at time $t$, the \pushforward operator is first recursively applied to the \tree to propagate the states up to the leaf nodes. Then, the subtask RMPs are evaluated on the leaf nodes and combined recursively  backward along the \tree by the \pullback operator. The \resolve operator is finally applied on the root node $\rtt$ to compute the desired acceleration $\pi(\q,\qd)=\ab_{\rtt}$.

\begin{figure*}[ht]
    \centering
    \begin{subfigure}[b]{0.28\linewidth}
        \centering
        \includegraphics[width=\columnwidth]{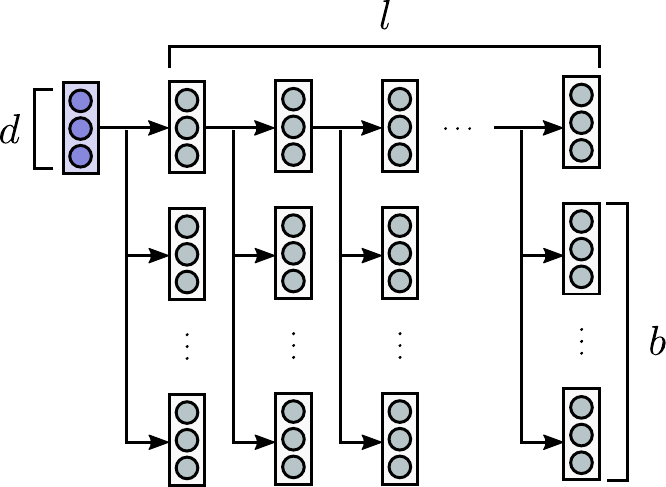}
        \caption{task map structure}
    \end{subfigure}
    \quad
    \begin{subfigure}[b]{0.2\linewidth}
        \centering
        \includegraphics[width=\columnwidth]{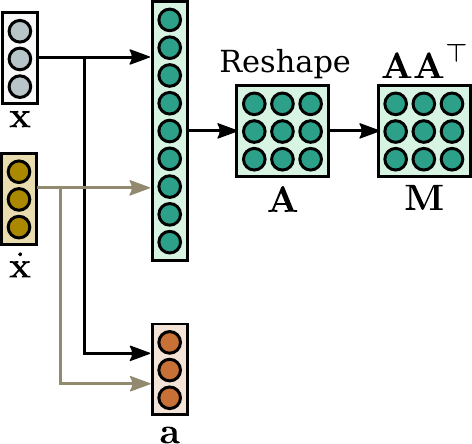}
        \caption{Policy structure}
    \end{subfigure}
    \quad
    \begin{subfigure}[b]{0.32\linewidth}
        \centering
        \includegraphics[width=\columnwidth]{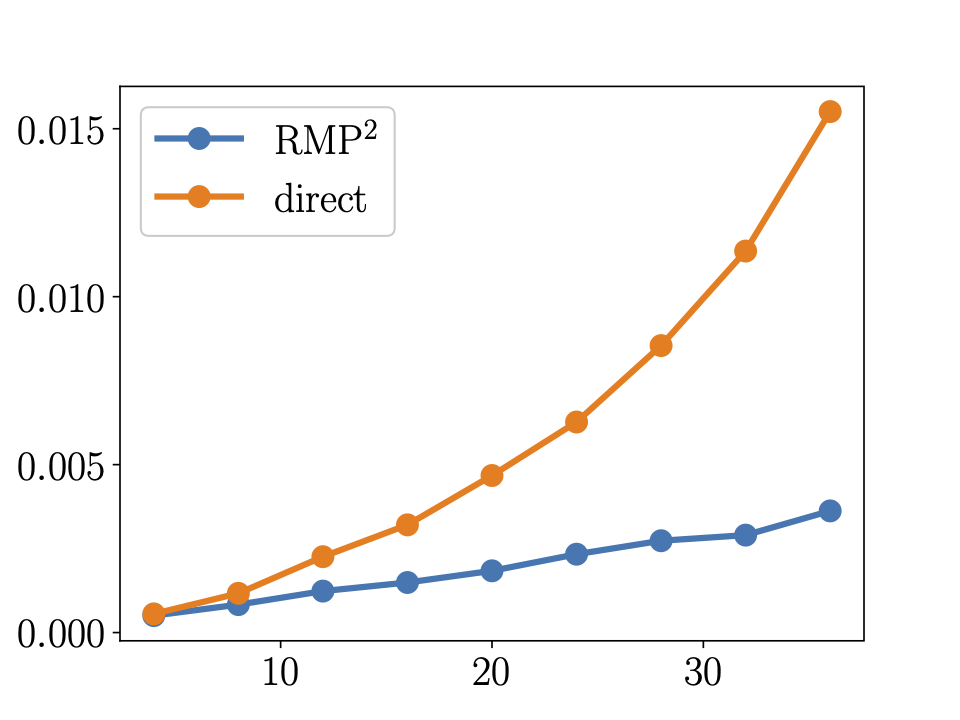}
        \vspace{-8mm}
        \caption{Computation time}
    \end{subfigure}
    \caption{(a) task map and (b) policy structure for benchmarking the complexity of the algorithms. (c) The computation time of \alg and the na\"ive implementation (direct) on chain-structured graphs with varying lengths. The results show that \alg has a linear time complexity whereas the na\"ive implementation (direct) has a super-linearly time complexity.}
    \label{fig:chain}
\end{figure*}

\section{Complexity Analysis of \alg}\label{sec:complexity}

In this section, we analyze the time and space complexities of \alg. We show that \alg has a time complexity of $O(Nbd^3)$ and a memory complexity of $O(Nd+Ld^2)$, where $N$ is the total number of nodes, $L$ is the number of leaf nodes, $b$ is the maximum branching factor, and $d$ is the maximum dimension of nodes
In comparison, we prove in Appendix~\ref{app:complexity} that the original \flow algorithm (\cref{alg:rmpflow}) by \citep{cheng2018rmpflow} has a time complexity of $O(Nbd^3)$ and a \emph{worse} space complexity of $O(Nd^2+Ld^2)$. 
Please see \cref{tb:theoretical comparison} for a summary.

Specifically, consider a directed-acyclic-graph-structured task map with $N$ nodes, where each node has dimension in $O(d)$ and has at most $b$ parents. We suppose that $L\le N$ nodes are leaf nodes, and that the \ad library is based on reverse-mode \AD.
We first analyze the complexity of task map evaluation and Jacobian-vector-product subroutine (\cref{alg:jvp}) based on reverse accumulation in preparation for the complexity analysis for \alg. 

\textbf{Task map evaluation: }For each node in the graph, the input and output dimensions are bounded by $O(bd)$ and $O(d)$, respectively. Hence, evaluating each node has a time complexity in $O(bd^2)$. Because each node is evaluated exactly once in in computing the full task map, the total time complexity of task map evaluation $O(Nbd^2)$. 
If the Gradient Oracle \texttt{gradient} will be called (as in \alg), the value of each node needs to be stored in preparation for the gradient computation. Overall this would require a space complexity in $O(Nd)$ to store the values in the entire graph.

\textbf{Jacobian-vector product with $L$ output nodes: }Suppose that the output of the graph in \cref{alg:jvp} is a collection of $L$ nodes in the graph. During reverse accumulation, the task map is first computed, which, based on the previous analysis, has time and space complexity of $O(Nbd^2)$ and $O(Nd)$, respectively. The dummy variable $\bm\lambda$ is of size $O(Ld)$ and computing the inner product $\bm\lambda^\t \vb$ requires $O(Ld)$ computation (i.e. it creats a new node of dimension $1$ with $2L$ parent nodes of dimension in $O(d)$). By the reverse-mode \AD assumption, the first backward pass on the graph (line~\ref{ln:auxiliary}) has time complexity of $O(Nbd^2+Ld)=O(Nbd^2)$ and space complexity of $O(Nd+Ld)=O(Nd)$~\cite{griewank2008evaluating}. The final backward pass (line~\ref{ln:jvp}) is on a graph of size $O(N)$, as the first backward pass creates additional $O(N)$ nodes. With a similar analysis, the second backward pass have time complexity of $O(Nbd^2 + Ld)=O(Nbd^2)$ and space complexity of $O(Nd+Ld)=O(Nd)$. Therefore, the time and space complexity of~\cref{alg:jvp} is $O(Nbd^2)$ and $O(Nd)$, respectively. 

\textbf{Forward pass: }
The complexity of line \ref{ln:taskmap}--\ref{ln:curvature} in~\cref{alg:new} follow the precursor analyses above. Here the computation graph is always of size $O(N)$ (the original task map is in $O(N)$ and each call of Gradient Oracle \texttt{gradient} in the Jacobian-vector-product subroutine \texttt{jvp} creates additional $O(N)$ nodes in the computation graph). By previous analysis, the time and space complexity of the forward pass are $O(Nbd^2)$ and $O(Nd)$, respectively. 

\textbf{Leaf evaluation: }
Assume, for each leaf node, $O(d^3)$ computation is needed for evaluating the weight matrix and $O(d^2)$ for acceleration in an RMP. The leaf evaluation step then have time complexity of $O(Ld^3)$ and space complexity of $O(Ld^2+Ld)=O(Ld^2)$. 

\textbf{Backward pass: } 
By previous analysis, task map evaluation requires $O(Nbd^2)$ computation and $O(Nd)$ space. The vector-matrix-vector product for computing auxiliary variables $r$ and $s$ has time complexity of $O(Ld^2)$. To compute the metric at root, $\M_\rtt$, the first backward pass, $\gradient{r}{\q}$ needs $O(Nbd^2)$ time and $O(Nd)$ space as it operates on a graph of size $O(N)$ where the number of parents of each node is in $O(b)$\footnote{Except the final node aggregating $L$ outputs. However, it does not change the complexity as it adds a complexity in $O(Ld^2) < O(Nbd^2)$.}. The \texttt{jacobian} operator in line~\ref{ln:backward2} is done by $O(d)$ sequential calls of the Gradient Oracle \texttt{gradient}. Hence, it has a time complexity of $O(Nbd^3)$ and a space complexity of $O(Nd)$. (Because we are not taking further derivatives, the values of the new graphs created in calling the \texttt{jacobian} operator do not need to be stored.)
With a similar analysis, computing $\f_\rtt$ requires $O(Nbd^3)$ computation and $O(Nd)$ space complexities.

\textbf{Resolve: }The matrix inversion has time complexity of $O(d^3)$ and space complexity of $O(d^2)$. 

In summary, the time complexity of \alg is $O(Nbd^2+Ld^3+Nbd^3+d^3)={O(Nbd^3)}$ and the space complexity is $O(Nd+Ld^2)$. 

\section{Complexity of \flow}\label{app:complexity}

First we need to convert a graph with $O(b)$ parent nodes into a tree. This can be done by creating meta nodes that merge all the parents of a node into a single parent node; inside the mega node, each component is computed independently. Therefore, for a fair comparison, in the following analysis, we shall assume that in the tree version evaluating each node would need a time complexity in $O(bd^2)$. We suppose the space complexity to store all the nodes is still in $O(Nd)$ because the duplicated information resulting from the creation of the meta nodes can be handled by sharing the same memory reference in a proper implementation~\citep{cheng2018rmpflow}.

\textbf{Forward pass: }
During the forward pass, similar to the reverse accumulation analysis, $O(bd^2)$ per node is needed for \texttt{pushforward}, i.e. computing the pushforward velocity through reverse accumulation, yielding a time complexity of $O(Nbd^2)$. The space complexity is $O(Nd)$ for storing the state at every node.

\textbf{Leaf evaluation: }Same as \alg. 

\textbf{Backward pass: }Computing the metric in \eqref{eq:pullback} requires $O(bd^3)$ computation per node, yielding time complexity of $O(Nbd^3)$ and space complexity of $O(Nd^2)$. The curvature term $\Jd_{\vtt_m;\utt}\xd$ can be computed through reverse accumulation similar to \alg, which has time complexity of $O(bd^2)$ per node. The matrix-vector products to compute the force requires $O(bd^2)$ space and computation for each node. 

\textbf{Resolve: }Same as \alg.

Thus, the time complexity of \flow is $O(Nbd^2+Ld^3+Nbd^3+Nbd^2+d^3)={O(Nbd^3)}$ and the space complexity is $O(Nd+Ld^2+Nd^2+bd^2+d^2)={O(Nd^2+Ld^2)}$. 

\begin{figure*}[ht]
    \centering
    \begin{subfigure}[b]{0.4\linewidth}
        \centering
        \includegraphics[width=\columnwidth]{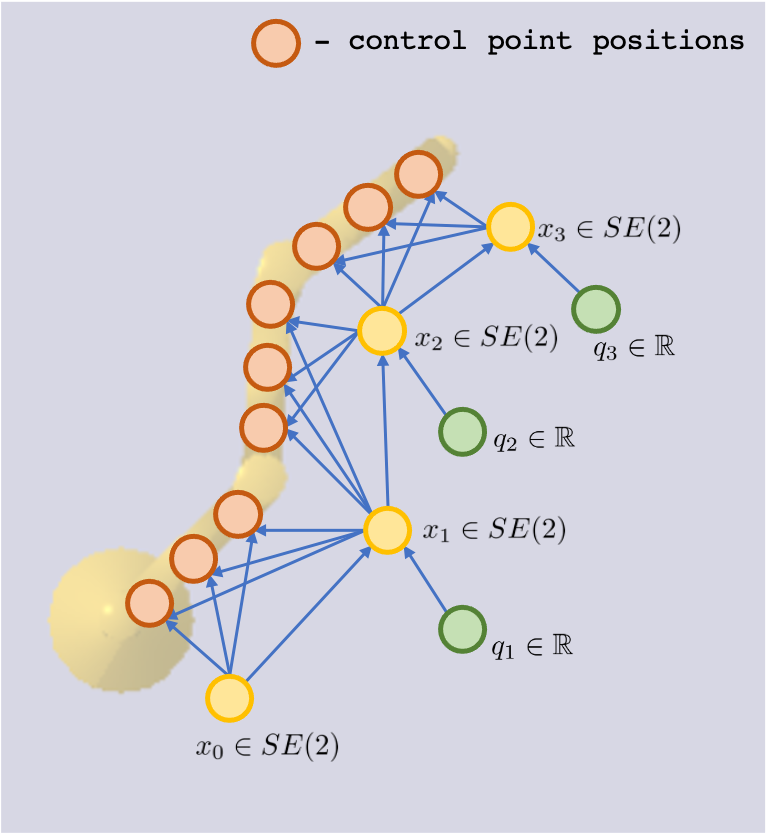}
        \caption{The computational graph used by \alg}
    \end{subfigure}
    \qquad
    \begin{subfigure}[b]{0.4\linewidth}
        \centering
        \includegraphics[width=\columnwidth]{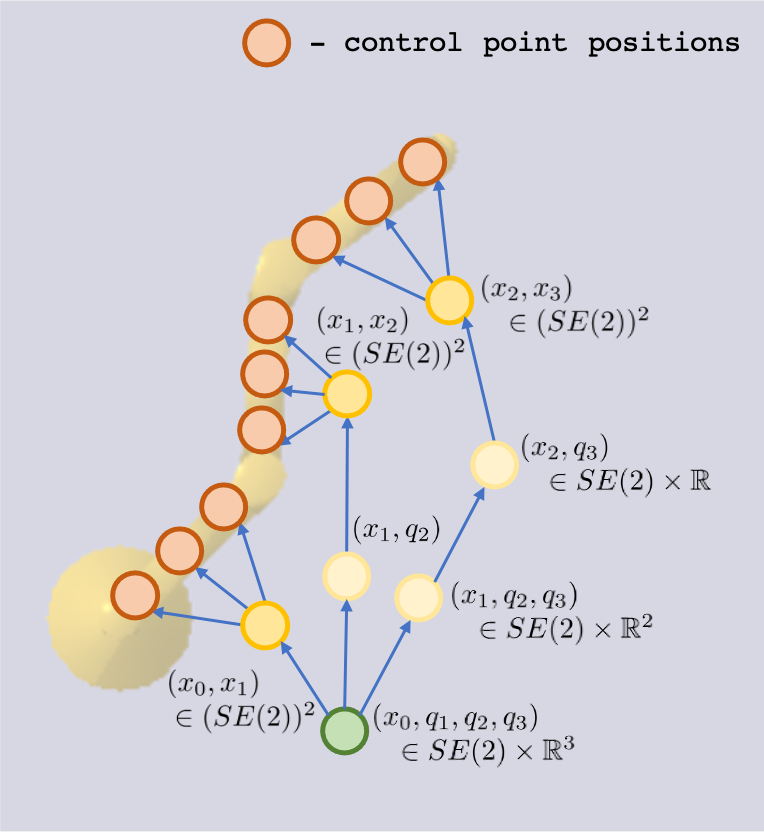}
        \caption{An example \tree which can be used by \flow}
    \end{subfigure}
    \caption{(a) The computation graph automatically built by \AD libraries when computing control point positions. \alg directly operates on this graph. (b) An example \tree constructed for the same problem under similar strategy as~\cite[Appendix D]{cheng2018rmpflow}. It introduces intermediate high-dimensional nodes as well as redundant computation. In both figures, $q_i$ is the joint angle of the $i$-th joint, $x_i$ is the pose (position and orientation) of the $i$-th link, and $x_0$ is the base pose. }
    \label{fig:tree}
\end{figure*}

\section{Experimental Validation of Time Complexity}
We validate the time complexity analysis for \alg and the na\"ive implementation (\cref{alg:direct,alg:new}, respectively). In particular, we are interested in how the two algorithms scale with respect to the number of nodes in the graph.

We consider a directed chain-like graph of length $l$: Each node on the chain is connected to $b$ leaf nodes; every node in the graph is of dimension $d$. Overall such a graph has $1+(b+1)l$ nodes, where $bl$ nodes are leaf nodes. The task map structure for the time complexity experiment is shown in~\cref{fig:chain}. 
In the experiment, we vary the length of the chain while fixing the branching factor\footnote{The branching factor and the chain length have similar effect to the size of the graph and hence only the effect of chain length is evaluated.} and the dimension of the nodes ($b=d=3$). The map associated with each edge is implemented as a single-layer neural network with \texttt{tanh} activation function. Both algorithms as well as the graph structure are implemented in TensorFlow~\cite{tensorflow2015-whitepaper}. We consider chain length in  $[4, 8,12, \ldots, 36]$. 

\cref{fig:chain}(c) shows the computation time of the two algorithms. The computation time of \alg increases linearly with respect to the size of the graph whereas the na\"ive implementation (direct) suffers from a super-linear growth.
The computation time reported in \cref{fig:chain}(c) is the average over $1000$ runs of the algorithms with random input on a {static} computational graph in  TensorFlow~\citep{tensorflow2015-whitepaper}. 
The \emph{constant} time required to compile the static computational graph is not included in the reported average computation time.

\section{\alg versus \flow: A Case Study}\label{sec:interface_case_study}

In this appendix, we demonstrate how \alg provides a more convenient interface for the user. Consider the planar three-link robot from the experiment section~(\cref{fig:franka}). Assume that the subtask spaces are with the positions of control points along the robot arm. In~\cref{fig:tree}, for example, there are $9$ spaces, each corresponding to the position of one control point on the robot. This type of subtask space is useful for specifying behaviors such as collision avoidance, where we need each control point to avoid collision with obstacles in the environment. 

Intuitively, to compute the control point positions along the robot, we can first compute the pose of each link, $\{x_i\}_{i=1}^3$, through the kinematic chain of the robot, where $x_i\in SE(2)$ denotes the position and orientation of the $i$-th link. Then, the control point positions can be obtained through interpolating the positions of any two adjacent links. \cref{fig:tree}(a) shows the computational graph that is automatically built through the above computation, where green nodes denote joint angles, yellow nodes denote link poses, and orange nodes denote control point positions. As is introduced in~\cref{sec:method}, \alg directly uses the computational graph as the core data structure and compute the policy through calling the Gradient Oracles provided by \AD libraries. 

In contrast, \flow (see Appendix~\ref{sec:rmpflow_alg}) requires the user to specify a tree data structure called the \tree. In the \tree structure, it is required that the states in the parent node contain sufficient information for computing the child node states (see the \pullback operator in Appendix~\ref{sec:rmpflow_alg}). Here we use a \tree structure (\cref{fig:tree}(b)) similar to what is introduced in \cite[Appendix D]{cheng2018rmpflow}. The root node includes $x_0$, the pose of the base, as well as all joint angles $\{q_i\}_{i=1}^3$ as we need all these quantities to compute the control point positions. Then, the \tree branches out to compute the control points on each link. To compute the control point positions for the first link, we need the poses\footnote{Only the positions are sufficient. However, we use poses here to make the notation more compact. } of the base link and link $1$, $(x_0,x_1)\in(SE(2))^2$, similarly for the other two links. This gives us a \tree structure shown in~\cref{fig:tree}(b). Note that computationally, to compute the poses of link $3$, for example, we need to compute the poses for all previous links, as is shown in the light yellow nodes on the path for the second and third links. Therefore, not only does the construction of \tree cost additional effort, it also introduces nodes with high dimensions (e.g. $SE(2)\times \R^2$) and redundant computation (the forward mapping for the first link is computed by all three branches) under less careful design choices. These, in turn, impair the computational efficiency of \flow as the time complexity is a function of node dimension and node number. 

Moreover, \alg allows us to specify complicated task maps more easily. For example, if one would like to additionally consider self-collision avoidance (even though it is not relevant to the planar three-link robot). For \alg, one can directly compute the distance between any two control points (perhaps from different links), creating nodes which are child nodes to pairs of control point position nodes. However, \flow requires the user to redesign the \tree structure entirely, creating even more intermediate nodes due to the limitation of the tree structure.

\end{document}